\documentclass[letterpaper, 10 pt, conference]{ieeeconf}  % Comment this line out if you need a4paper

\IEEEoverridecommandlockouts                              % This command is only needed if 
\usepackage{graphics} % for pdf, bitmapped graphics files
\usepackage{epsfig} % for postscript graphics files
\usepackage{times} % assumes new font selection scheme installed
\usepackage{amsmath} % assumes amsmath package installed
\usepackage{amssymb}  % assumes amsmath package installed
\usepackage{caption}
\usepackage{lipsum}
\usepackage{float}

\usepackage{algorithm}
\usepackage{algorithmic}
\usepackage{color}
\usepackage{booktabs} % 导入三
\usepackage{subfigure}
\usepackage{setspace}
\usepackage{bm}
\usepackage{color}
\usepackage{multirow}

\begin{document}
\title{\LARGE \bf
Decentralized Motor Skill Learning for Complex Robotic Systems
}

\author{Yanjiang Guo$^{1*}$, Zheyuan Jiang$^{1*}$, Yen-Jen Wang$^{1}$, Jingyue Gao$^{1}$, Jianyu Chen$^{1,2}$
% <-this % stops a space
\thanks{$^{*}$ Equal contribution.}
\thanks{$^{1}$ Yanjiang Guo, Zheyuan Jiang, Yen-Jen Wang, Jingyue Gao, Jianyu Chen are with Institute of Interdisciplinary Information Sciences, Tsinghua University, Beijing, China. {\tt\small guoyj22@mails.tsinghua.edu.cn}}
\thanks{$^{2}$ Jianyu Chen is also with Shanghai Qi Zhi Institute, Shanghai, China (Corresponding author). {\tt\small jianyuchen@tsinghua.edu.cn}}
% <-this % stops a space
% \thanks{$^{1}$Albert Author is with Faculty of Electrical Engineering, Mathematics and Computer Science,
%         University of Twente, 7500 AE Enschede, The Netherlands
%         {\tt\small albert.author@papercept.net}}%
% \thanks{$^{2}$Bernard D. Researcheris with the Department of Electrical Engineering, Wright State University,
%         Dayton, OH 45435, USA
%         {\tt\small b.d.researcher@ieee.org}}%
}

% \twocolumn[{
% \renewcommand\twocolumn[1][]{#1}
% \maketitle
% \begin{center}
%     \vspace{-6mm}
%     \captionsetup{type=figure}
%     \includegraphics[width=1.0\textwidth]{figures/pic1_v2_clip.pdf}
%     \captionof{figure}{Brief framework of \textbf{Di}stributed \textbf{mo}tor \textbf{s}kill learning (\textbf{DEMOS}). DEMOS automatically decouple some connections between different body parts while keeping the important cooperation, which results in decentralized policies that are more robust and transferable.}
% \end{center}
% }]
\maketitle
% \blfootnote{$^{*}$ Equal contribution.}
% \blfootnote{$^{1}$ Yanjiang Guo, Zheyuan Jiang, Yen-Jen Wang, Jingyue Gao, Jianyu Chen are with Institute of Interdisciplinary Information Sciences, Tsinghua University, Beijing, China. {\tt\small guoyj22@mails.tsinghua.edu.cn}}
% \blfootnote{$^{2}$ Jianyu Chen is also with Shanghai Qi Zhi Institute, Shanghai, China (Corresponding author). {\tt\small jianyuchen@tsinghua.edu.cn}}%

\thispagestyle{empty}
\pagestyle{empty}

%%%%%%%%%%%%%%%%%%%%%%%%%%%%%%%%%%%%%%%%%%%%%%%%%%%%%%%%%%%%%%%%%%%%%%%%%%%%%%%%

\begin{abstract}

% Inspired by biological principles, we have developed a decentralized control system utilizing a reinforcement learning framework. This approach emphasizes the independence of each body part, allowing for the automatic identification of discretely controllable segments while preserving necessary cooperation between them. Our design divides the robot into distinct modules based on function, each with its own control network that takes local information as input and outputs motor actions globally. However, actions that fall outside the scope of the current module are kept close to zero. The final outcome is achieved by summing the outputs of all networks. Our proposed method has been tested and shown to deliver more robust performance in both quadruped and biped robots, and has the added advantage of allowing individual module skills to be transferred to new tasks, enabling efficient completion of demanding tasks.

Reinforcement learning (RL) has achieved remarkable success in complex robotic systems (eg. quadruped locomotion). In previous works, the RL-based controller was typically implemented as a single neural network with concatenated observation input. However, the corresponding learned policy is highly task-specific. Since all motors are controlled in a centralized way, out-of-distribution local observations can impact global motors through the single coupled neural network policy. In contrast, animals and humans can control their limbs separately. Inspired by this biological phenomenon, we propose a Decentralized motor skill (DEMOS) learning algorithm to automatically discover motor groups that can be decoupled from each other while preserving essential connections and then learn a decentralized motor control policy. Our method improves the robustness and generalization of the policy without sacrificing performance. Experiments on quadruped and humanoid robots demonstrate that the learned policy is robust against local motor malfunctions and can be transferred to new tasks.
% results in more robust policies, and some modules can even work independently and be transferred to new tasks.

% decouple robot modules that will not sacrifice performance and only keep essential connections. 

\end{abstract}

%%%%%%%%%%%%%%%%%%%%%%%%%%%%%%%%%%%%%%%%%%%%%%%%%%%%%%%%%%%%%%%%%%%%%%%%%%%%%%%%
\section{INTRODUCTION}

Recently, many complex robotic systems have been developed and demonstrated. Quadruped robots, for instance, have shown their ability to traverse diverse challenging terrains \cite{hwangbo2019learning,lee2020learning,miki2022learning}. Furthermore, attaching an additional arm to a quadruped robot allows it to perform various manipulation tasks \cite{fu2022deep}. 
Some impressive biped and humanoid robots also show great improvements in their locomotion and manipulation skills \cite{kaneko2019humanoid,siekmann2021sim, kuindersma2016optimization}.
% Biped robots can also generate a variety of rich gaits \cite{siekmann2021sim}. Robots with more complex structures such as humanoid robots are also developing rapidly \cite{kaneko2019humanoid,siekmann2021sim}.
However, controlling these complex robotic systems is a challenging task. Due to complex robot dynamics and environments, model-based control methods might fail when the model largely deviates from the ground truth. 
Reinforcement learning provides an alternative way to learn control policies without the need to model the dynamics and environments. Such mechanism helps improve robustness and generalization \cite{hwangbo2019learning,lee2020learning,miki2022learning,kumar2021rma}. % Reinforcement learning methods have addressed these challenges and made great progress in the quadruped and biped locomotion tasks, with good generalization performance and fast response \cite{miki2022learning}.

\begin{figure}[ht]
    \centering
    \includegraphics[width=0.49\textwidth]{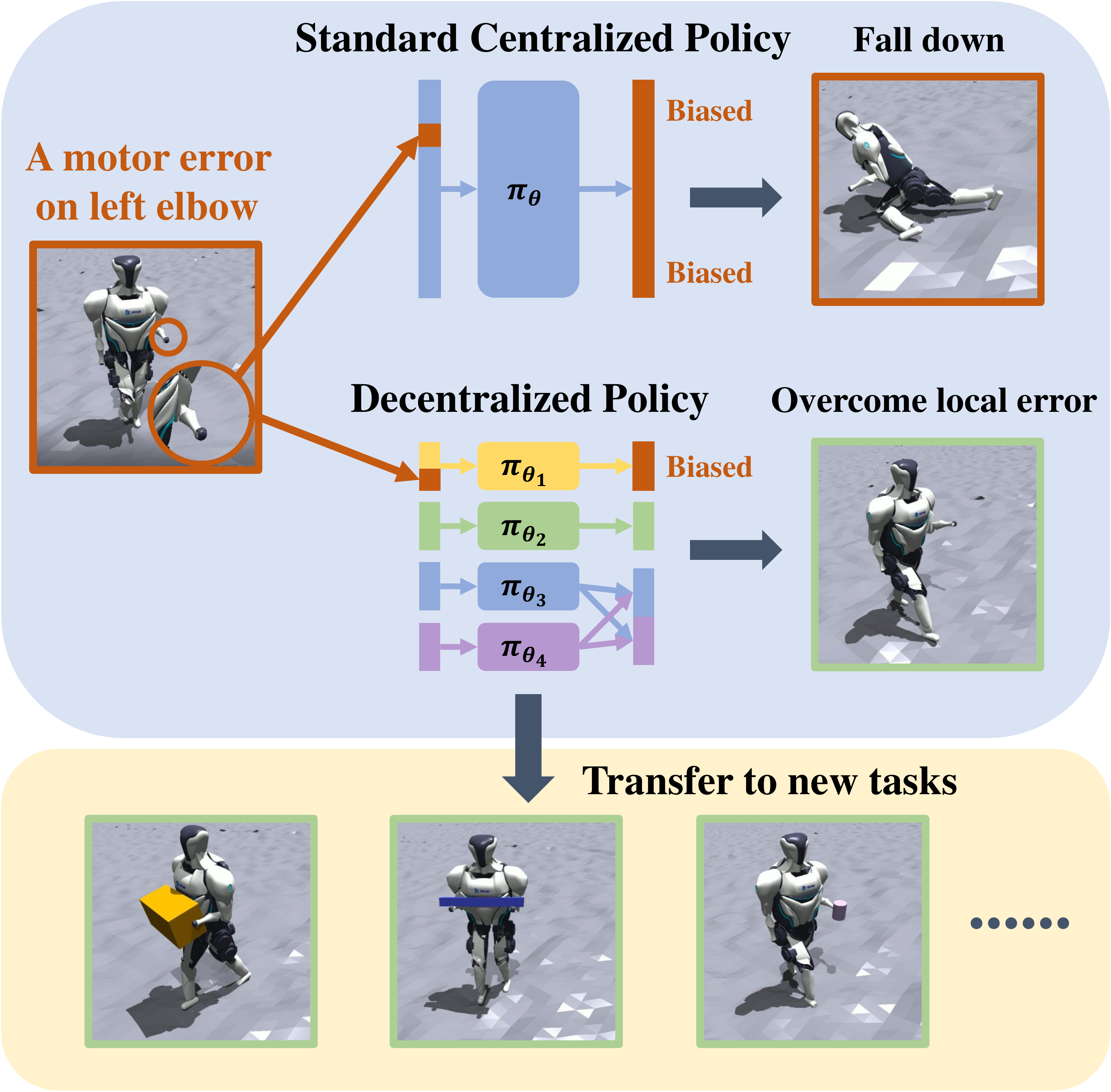}
    \caption{Illustration of our motivation. Standard centralized policies are highly task-specific and local motor malfunction can influence actions for all motors. \textbf{DEMOS} can automatically decouple some motor groups, making the policy more robust and transferable.}
    \label{motivation}
    \vspace{-5mm}
\end{figure}

A typical reinforcement learning controller for robot motor control is implemented as a centralized neural network policy. The input to the policy is some important observations, such as root pose, joint pose, joint velocity, etc. All these observations are concatenated into a single vector, which is directly fed to the neural network, and then the neural network will directly output the actions such as target poses or the torques of the joints. Such an approach has some important drawbacks: 1) The policy lacks robustness. Since all motors are controlled in a centralized way, out-of-distribution local observations can impact global motors through the single coupled neural network policy, as shown in Figure \ref{motivation}. 2) The converged policy is highly specific to the training task and will be failed when transferred to new tasks.
In order to address the challenges of controlling complex robotic systems, previous work has employed a decentralized approach. De et al. manually partitioned the robot motors into multiple groups and treated each group as a single agent \cite{de2020deep}. These agents used their local observations to control their corresponding motors. However, this manual partitioning could potentially disrupt the coordination between motor groups and sacrifice overall performance. In contrast, animals and humans have adopted a gated mechanism \cite{leiras2022brainstem,lindsay2020attention} that automatically decides which modules need to cooperate with each other and which can function independently. For instance, humans can still walk naturally with a broken arm, as it is independent from their legs, but humans would struggle to walk with a broken leg, which requires coordination with the other leg.

% automatically discover motor groups that can be decoupled from each other while preserving essential connections and then learn a decentralized motor control policy. Our method improves the robustness and generalization of the policy without sacrificing performance. Experiments on quadruped and humanoid robots demonstrate that the learned policy is robust against local motor malfunctions and can be transferred to new tasks.

Inspired by this biological phenomenon, we propose a \textbf{De}centralized \textbf{mo}tor \textbf{s}kill learning (\textbf{DEMOS}) method under the RL framework. 
DEMOS automatically discovers motor groups that can be decoupled from each other while preserving essential connections and then learn a decentralized motor control policy, as shown in Figure 1. 
Specifically, DEMOS first divide robots into potentially decoupled motor groups and then introduces a decentralized objective to encourage each group to influence others as little as possible. Finally, DEMOS decouple groups with weak connections and obtain a decentralized policy. Further experiments on quadruped robots and humanoid robots show the robustness and generalization of the learned decentralized policy.

% This approach emphasizes the independence of each body part, allowing for the automatic identification of discretely controllable segments while preserving necessary cooperation between them. Our design divides the robot into distinct modules based on function, each with its own control network that takes local information as input and outputs motor actions globally. However, actions that fall outside the scope of the current module are kept close to zero. The final outcome is achieved by summing the outputs of all networks. Our proposed method has been tested and shown to deliver more robust performance in both quadruped and biped robots and has the added advantage of allowing individual module skills to be transferred to new tasks, enabling efficient completion of demanding tasks.

Our main contributions can be summarized as follows:
\begin{itemize}
    \item We propose a distributed motor skill learning (DEMOS) method inspired by biological phenomenons, which can automatically decouple motor groups for better robustness and generalization.
    \item We apply the DEMOS algorithm to both a quadruped robot and a humanoid robot and demonstrate that DEMOS can learn a robust policy against motor malfunctions.
    \item We demonstrate that the learned decentralized policy can be directly transferred to new tasks on the humanoid robot.
\end{itemize}

\begin{figure}[ht]
    \centering
    \includegraphics[width=0.49\textwidth]{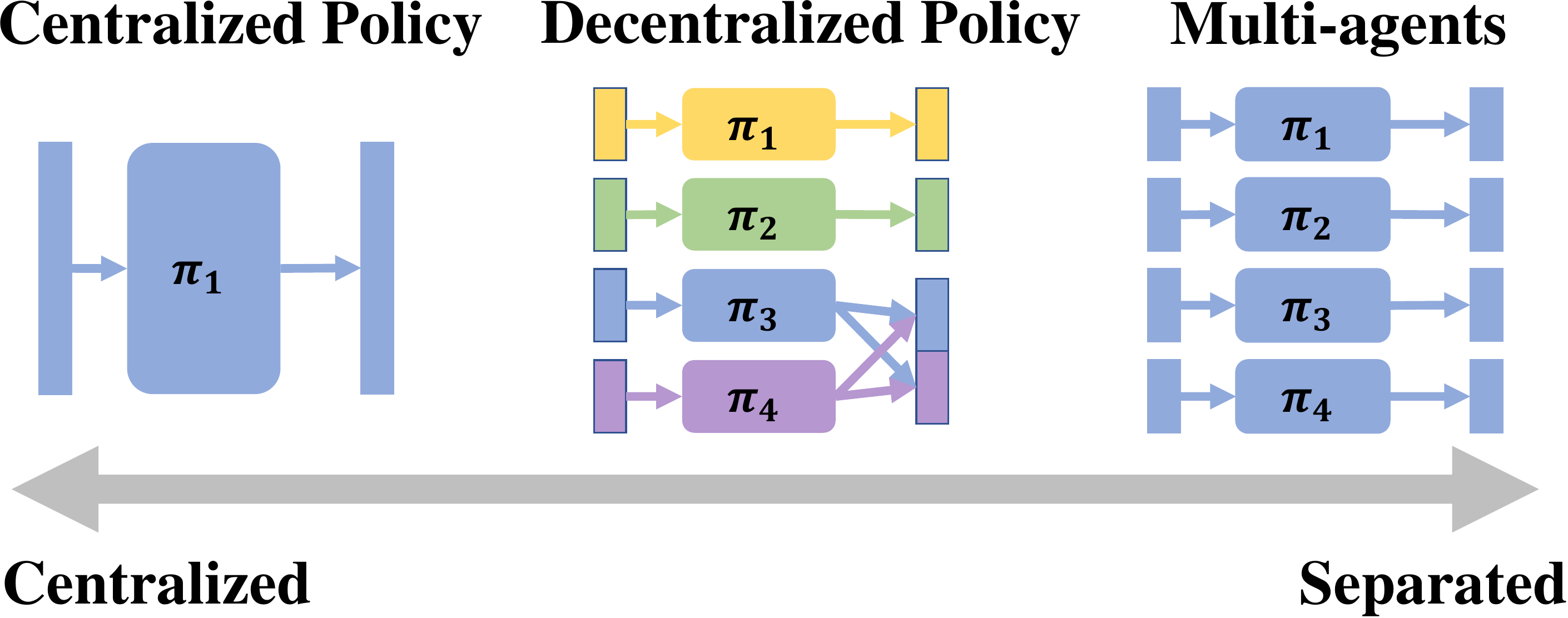}
    \caption{Single neural network policy is highly centralized and task-specific while manually dividing the robot into multi-agents may disrupt useful coordination. \textbf{DEMOS} partially decouples modules and keeps essential connections during skill learning.}
    \label{figcompare}
    % \vspace{-4mm}
\end{figure}

\section{RELATED WORKS}
\subsection{Reinforcement Learning for Legged Robots}
% Legged robots possess the ability to navigate a variety of terrains that are inaccessible to traditional wheeled or tracked robots. Quadruped robots can learn to locomote on various terrains through reinforcement learning, employing manually designed rewards.

% Legged robots have the potential to traverse various terrains that are unreachable by wheeled robots. Through reinforcement learning with manually designed rewards, quadruped robots can learn to locomote on various terrains. Furthermore, through imitation learning \cite{escontrela2022adversarial, peng2018deepmimic, peng2020learning} on animal mocap data, quadruped robots can generate more diverse and natural gaits. To perform additional manipulation tasks, a robotic arm can be attached to the quadruped robot. Ma et al.\cite{ma2022combining} use reinforcement learning for the locomotion task and use a model-based controller for robotic arm manipulation; Fu et al.\cite{fu2022deep} use reinforcement learning to jointly optimize the control of quadruped and attached arm. 
% Besides quadruped robots, reinforcement learning with hand-designed rewards is also used for biped and humanoid robots, which generate a wide variety of robust gaits \cite{siekmann2021sim}. \textbf{digit}
% Humanoid robots are more complex systems, and reinforcement learning that relies purely on hand-designed reward functions is difficult to produce natural behaviors in humanoid robots \cite{merel2017learning}. Therefore, human mocap data is incorporated into the training process for guidance or regularization \cite{peng2018deepmimic}.
Legged robots have the potential to traverse various terrains that are inaccessible by wheeled robots. Quadruped robots can learn to locomote on diverse terrains through reinforcement learning with manually designed rewards. Additionally, quadruped robots can generate more diverse and natural gaits through imitation learning on animal mocap data, as demonstrated by \cite{escontrela2022adversarial,peng2018deepmimic, peng2020learning}.
To perform additional manipulation tasks, a robotic arm can be attached to the quadruped robot. Ma et al. used reinforcement learning for the locomotion task and a model-based controller for robotic arm manipulation \color{black}\cite{ma2022combining}\color{black}. Meanwhile, Fu et al. used reinforcement learning to jointly optimize the control of the quadruped and attached arm \color{black}\cite{fu2022deep}\color{black}.
Aside from quadruped robots, reinforcement learning with hand-designed rewards is also applied to biped and humanoid robots, resulting in a wide variety of robust gaits \cite{siekmann2021sim,radosavovic2023learning}.
\subsection{Decentralized Robot Motor Control}
% The most common way to control a robot system through reinforcement learning is to concatenate all proprioceptive variables (for example, joint position or speed) together, and use it as the input of a single neural network which then outputs the target position or the torques for all motors \cite{hwangbo2019learning,lee2020learning,miki2022learning}. Such a centralized approach can achieve better performance on the training task, while the converged policy is highly specific for the task and always needs extra efforts for new tasks \cite{finn2017model,yin2019meta}. Moreover, all joints are coupled together in a centralized neural network policy and cannot work independently, which reduces robustness and interpretability.

The most common reinforcement learning controller is a single centralized neural network with concatenated observation input. This centralized policy is highly specific to the training task. Moreover, all robot modules are coupled together and can not function independently.
% the RL-based controller was typically implemented as a single neural network with concatenated observation input. However, the corresponding learned policy is highly task-specific. Since all motors are controlled in a centralized way, out-of-distribution local observations can impact global motors through the single coupled neural network policy.

% To address these limitations, many works have adopted a decentralized approach that divides the robot into modules and utilizes multiple policies to control different modules \cite{whitman2021learning,wang2018nervenet}. Ma et al. separate the quadruped-arm robot into the locomotion part and manipulation part and control them through a combination of data-driven and model-based methods \cite{ma2022combining}. The most extreme form of decentralization would be to treat each joint as an individual module and only receive information from neighboring joints \cite{huang2020one}. From another perspective, decentralized control can be viewed as a multi-agent collaboration problem \cite{de2020deep,peng2021facmac}, where each module is considered as an agent that observes local information and controls its own motors. However, all these decentralized control methods mentioned above typically require manual division before training, which may impact the overall performance. In contrast, our method automatically decouples motor groups while still preserving the connection between groups that are essential to the whole body performance, thus avoiding compromise on overall performance. Comparison can be found in Figure \ref{figcompare}.

To address these limitations, several works have adopted a decentralized approach that partitions the robot into modules and utilizes multiple policies to control different modules \cite{whitman2021learning,wang2018nervenet}. For instance, Ma et al. \cite{ma2022combining} separate the quadruped-arm robot into the locomotion part and the manipulation part, controlling them via a combination of data-driven and model-based methods. Decentralized control can be viewed as a multi-agent collaboration problem \cite{de2020deep,peng2021facmac}, where each module is treated as an agent that observes local information and controls its own motors. Huang et al. \cite{huang2020one} propose the most extreme form of decentralization, where each joint is treated as a separate module, and only receives information from neighboring joints. However, all the above-mentioned decentralized control methods typically require manual partitioning before training, which can affect the overall performance.
In contrast, our method automatically decouples motor groups while still maintaining the essential connections between groups, which are necessary for whole-body performance, thus avoiding any compromise on overall performance. A comparison of our method with other approaches can be found in Figure \ref{figcompare}.

\begin{figure*}[ht]
    % \vspace{-4mm}
    \centering
    \includegraphics[width=1.0\textwidth]{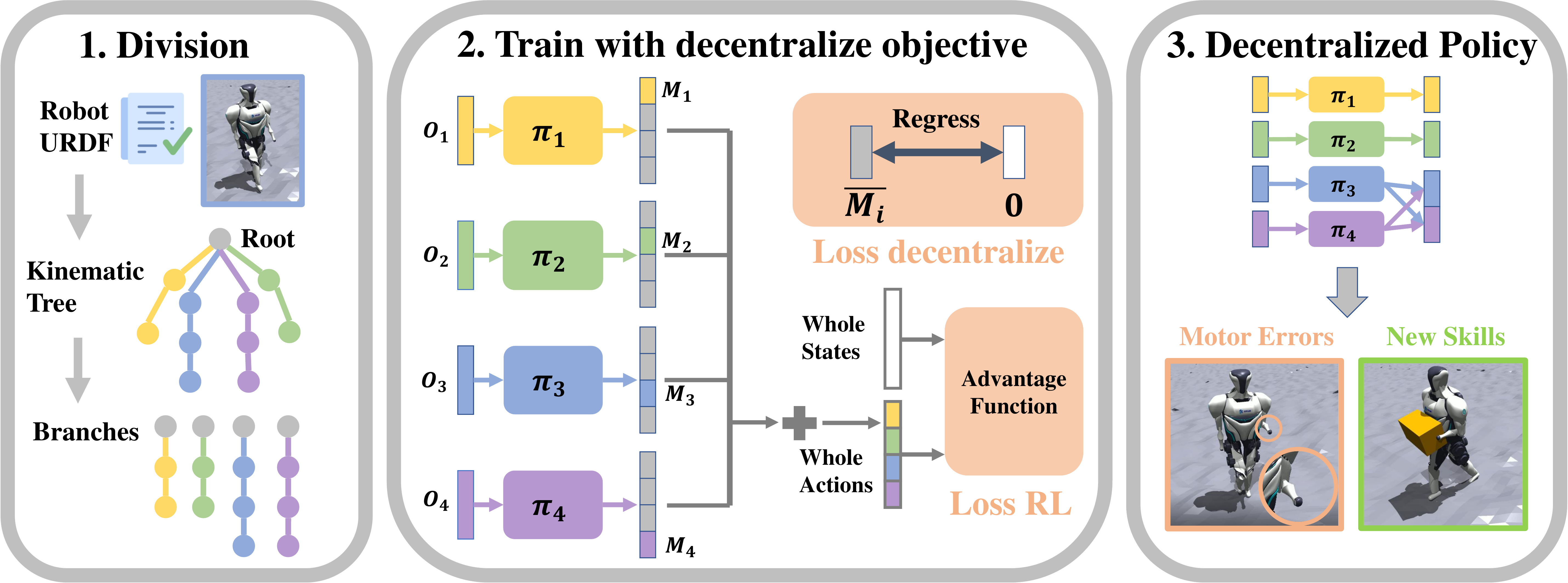}
    \caption{Three steps of the decentralized motor skill learning pipeline: 1) Divide the robot into branches according to the kinematic tree described in its URDF. 2) $O_i$ and $M_i$ are local observations and actions for branch $B_i$. Actions out of $M_i$ (colored grey) are encouraged to maintain zero. 3) Decouple modules with weak connections and obtain decentralized policies.}
    \vspace{-5mm}
    \label{figmethods}
\end{figure*}

\section{PRELIMINARY}

Instead of controlling the robot with a single neural network policy, we divide the robot into $n$ potentially decoupled modules and control them in a decentralized way. This setting can be constructed under the decentralized partially observable Markov decision processes (Dec-POMDP) \cite{oliehoek2016concise} defined by $\left\langle \mathbb{D}, \mathbb{S},\mathbb{A},\mathbb{O},O,P,R,\gamma \right\rangle$. $\mathbb{D}$ is the set of $n$ modules and $\mathbb{S}$ is the state space. $\mathbb{O}_i$ is the set of observations available to the $i^{th}$ module with elements $o_i = O(s;i)$. $\mathbb{M}_i$ is the motor group on $i^{th}$ module. Straightforwardly, the global observation space $\mathbb{O}$ and global action space $\mathbb{A}$ are the aggregation of module's sub-spaces:
\begin{equation}\label{global}
    \mathbb{O} = \mathbb{O}_1\cup \mathbb{O}_2\cup...\cup \mathbb{O}_n  \quad \mathbb{A} = \mathbb{M}_1\cup \mathbb{M}_2\cup...\cup \mathbb{M}_n
\end{equation}
Specifically in our setting, the $i^{th}$ module holds a policy $\pi_{\theta_i}$ which \textbf{inputs local observations \bm{$o_i$} but outputs global actions \bm{$a_i$}}. This indicates that each module's local observations can contribute to whole-body actions, and the final action $a$ is the sum of all module actions:
\begin{equation}
    a=a_1+a_2+...+a_n.
\end{equation}
$P(s'|s,a)$ denotes the transition probability from $s$ to $s'$ given the whole joint action $a$. $R(s,a)$ denotes the shared reward function and $\gamma$ is the discount factor. All modules cooperate with each other to maximize the discounted accumulated reward:
\begin{equation}
    J(\theta_1,\theta_2,..,\theta_n)=\mathbb{E}_{a^t,s^t} \left[ \sum_t \gamma^t R(s^t,a^t) \right]
\end{equation}

After initialization, all modules remain coupled with each other since all local observations contribute to global actions. During the training procedure, our proposed method discourages local observations $o_i$ from contributing to motors outside of $M_i$, and only keeps essential connections between modules. The detailed pipeline can be found in Section \ref{secmethod} or Figure \ref{figmethods}.

Our setting differs significantly from the settings that treat decentralized control problems as multi-agent problems. 
In the multi-agent setting, each agent observes locally and only controls its own motor group, which may disrupt coordination between motor groups. Moreover, since different modules are not homogeneous agents (i.e. agents have different observation spaces), some important technical designs such as policy parameter sharing \cite{terry2020revisiting,christianos2021scaling} can not be adopted.

%%%%%%%%%%%%%%%%%%%%%%%%%%%%%%%%%%%%%%%%%%%%%%%%%%%%%%%%%%%%%%%%%%%%%%%%%%%%%%%%%%%%%%%%%%%%%%%%%%%%%%%%%%%%%%%%%%%%%%%%%%%%%%%%%%%%%%%%%%%%%%%%%%%%%%%%%%%%%%%%%%%%%%%%%%%%%%%%%%%%%%%%%%%%%%%%%%%%%%%%%%%%%%%%%%%%%%%%
\section{METHODS}\label{secmethod}
In this section, we introduce the proposed \textbf{De}centralized \textbf{mo}tor \textbf{s}kill learning (\textbf{DEMOS}) algorithm for complex robot systems. Our pipeline can be summarized in the following steps:
\begin{enumerate}
    \item Pre-training stage (Sec. \ref{divide}): Divide the robot into potentially decoupled modules according to the URDF file.
    \item Training stage (Sec. \ref{loss}): Introduce a decentralized control objective to diminish the connections between modules without sacrificing performance.
    \item Post-training stage (Sec. \ref{policy}): Obtain decentralized policy by decoupling modules with weak connections and keeping the essential connections.
\end{enumerate}

The overall algorithm can be found in algo. \ref{algo}.

\subsection{Division Rules}\label{divide}
We begin by constructing a kinematic tree from the corresponding Universal Robot Description Format file (URDF). In this context, a branch refers to a path within the tree that originates from the root node and terminates at a leaf node. Multiple branches are present in every tree structure. Conceptually, joints and links located on the same branch of the tree are inherently interconnected in a sequential manner, thereby directly influencing their kinematics and dynamics. On the other hand, joints located on distinct branches of the tree exhibit relatively weak connections, as their forces or torques are conveyed through the root node, and these influences can be reflected in root velocity, acceleration, and other related parameters.

\color{black}Based on the aforementioned analysis, the robot components are categorized into $n$ branches with each branch originating from the root node and extending to a leaf node. Joints that belong to the same branch form a group or branch denoted as $B_i$. For instance, in the case of a humanoid robot with four limbs, there will be four branches. It's worth mentioning that branches may overlap and the same joint or link can exist in multiple branches. This process is illustrated in Figure \ref{figmethods} (1). 

To represent the local information related to branch $B_i$, we employ the notation $\mathbb{O}_i$. This encompasses the local joint pose, joint velocity, root projected gravity, and central periodic clock signal. Similarly, $\mathbb{M}_i$ represents the local motor situated on branch $B_i$. Consequently, the global observation space $\mathbb{O}$ and global action space $\mathbb{A}$ can be obtained using Equation \ref{global}.\color{black}
Then we initiate policies $\pi_{\theta_i}(1\leq i\leq n)$ for branch $B_i$, which \textbf{input local observation} \bm{$o_i$} \textbf{and output global action} \bm{$a_i$}. All branches are coupled with each other after initialization since all branches contribute to global actions.
% Then, we initiate the local-input-global-output policy for each branch, keeping the possibility of inter-branch cooperation to avoid degradation of performance. Details can be seen on the left side of Figure \ref{figmethods}.

\subsection{Decentralized Control Objective}\label{loss}
During the training process, in addition to the reinforcement learning (RL) objective, we also introduce a decentralized control objective that encourages each branch to affect other branches as little as possible. We denote motors not on branch $B_i$ as a complementary set $\overline{\mathbb{M}_i}$, and we have the following relationships:
\begin{equation}
    \mathbb{A} = \mathbb{M}_i \cup  \overline{\mathbb{M}_i},\; 1\leq i\leq n
\end{equation}
We minimize influences between branches through regressing actions in $\overline{\mathbb{M}_i}$ to $0$ \color{black}through L-P norm. Formally, for  policies with global outputs and a sampled transition batch $D$ consists of tuples $(o_1,...,o_n,s,a)$: 
\begin{equation}\label{loss1}
\begin{split}
    % \mathcal{J}_{de} = -\frac{1}{|D|} \sum_{i=1}^{n}\left[\sum_{\substack{o_i\sim D\\m\in \overline{\mathbb{M}_i}}}\left(\pi_{\theta_i}(o_i)[m]-0\right)^p\right]^{1/p}
    \mathcal{J}_{de} = -\frac{1}{|D|} \sum_{o_i\sim D}\left[\sum_{m\in \overline{\mathbb{M}_i}}\left(\pi_{\theta_i}(o_i)[m]-0\right)^p\right]^{1/p}
\end{split}
\end{equation}\color{black}
$m$ is the motor sampled from $\overline{\mathbb{M}_i}$ and $\pi_{\theta_i}(o_i)[m]$ stands for one dimensional action that controls motor $m$. $s$ represents all state information in this paper, including all local joint states and root states.
% For branch $B_i$ with policy $\pi_{\theta_i}$, we encourage action dimensions for motors outside of the branch $B_i$ to be close to 0.

The reinforcement learning objective can be written as follows:
\vspace{-3mm}
\begin{equation}\label{loss2}
\begin{split}
    \mu = \sum_{i=1}^{n} \pi_{\theta_i}(o_i), \quad a \sim \mathcal{N}(\mu, \sigma^2)\\
    \mathcal{J}_{RL} = \frac{1}{|D|} \sum_{(o_1,.,o_n,s,a)\sim D}\log P(a|s) \cdot A(s,a)
\end{split}
\end{equation}
Here $\mathcal{N}(\mu, \sigma^2)$ stands for a diagonal Gaussian distribution with mean $\mu$ and standard deviation $\sigma$. $A(s,a)$ stands for the advantage function.

Finally, the overall objective function is composed of both the RL objective and the decentralized control objective with learnable parameters $\theta_1,\theta_2,...,\theta_n, \sigma$. The proportion of the two objectives is controlled by a hyper-parameter $\lambda$:
\begin{equation}\label{lossall}
\begin{split}
    \mathcal{J}(\theta_1,\theta_2...,\theta_n, \sigma) = \mathcal{J}_{RL} + \lambda \mathcal{J}_{de}
\end{split}
\end{equation}
% However, these two parts of objectives may conflict with each other. Reinforce learning may encourage branches to cooperate with each other to obtain higher performance, while the decentralized control objective dis-encourage. In practice, we choose a small $\lambda$ such as 0.01. The main objective is still to obtain high performance in the training task and we only encourage diminishing the connection between branches that will not sacrifice performance. 

However, these two objectives may conflict with each other. Reinforcing learning may encourage branches to cooperate with each other to achieve higher performance, while the decentralized control objective discourages such cooperation. In practice, a small value of lambda, such as 0.01, is typically chosen. The primary objective remains to achieve high performance in the training task, while simultaneously minimizing the connection between branches without compromising performance.

\begin{figure}[t]
\begin{algorithm}[H]
   \caption{\textbf{De}centralized \textbf{Mo}tor \textbf{S}kill Learning (\textbf{DEMOS})}
   \label{algo}
\begin{algorithmic}
   \STATE{\textcolor[RGB]{100,100,220}{\# Pre-training stage}}
   \STATE{Divide robot into $n$ branches $B_1\sim B_n$ with observation space $\mathbb{O}_1\sim \mathbb{O}_n$ and motor sets $\mathbb{M}_1\sim \mathbb{M}_n$. Global observation space $\mathbb{O}$ and global action space $\mathbb{A}$ are defined by equation \ref{global}.}
   \STATE{Initialize policies $\pi_{\theta_1}\sim \pi_{\theta_n}$ with local observations $o_1\sim o_n$ and global actions $a_1\sim a_n$.}
   
   \STATE{\textcolor[RGB]{100,100,220}{\# Training stage}}
   \STATE {\bfseries Input: }Number of episodes $n_e$, steps of policy updates per epoch $n_p$, policies $\pi_{\theta_1}\sim \pi_{\theta_n}$, replay buffer $D$, proportion coefficient $\lambda$.
   \FOR {$k$ {\bfseries in} \{0, 1, 2, ...,$n_e$\}}
   \STATE Interact with the environment $a=\sum_{i=1}^{n} a_n$, full fill the buffer $D$ with $(o_1,...o_n,s,a)$.
   \FOR {$i$ {\bfseries in} \{0, 1, 2, ...,$n_p$\}}
   \STATE Update $\theta_1\sim \theta_n$ with objective $\mathcal{J} = \mathcal{J}_{RL} + \lambda \mathcal{J}_{de}$ (\ref{lossall})
   \ENDFOR
   \ENDFOR
   
   \STATE{\textcolor[RGB]{100,100,220}{\#  Post-training stage}}
   \FOR {$1\leq i\leq n, 1\leq j\leq n, i\neq j$}
   \STATE{Calculate connection strength $C_{ij}$ between branches $B_i$ and $B_j$ with equation \ref{connection}.}
   \IF{$C_{ij}<\eta$}
   \STATE{Decouple influences from $B_i$ to $B_j$ with equation \ref{decouple}.}
   \ENDIF
   \ENDFOR
   \STATE{Obtain decentralized motor policy $\pi:\{\pi_{\theta_1},\pi_{\theta_2},...,\pi_{\theta_n}\}$.}
\end{algorithmic}
\end{algorithm}
\vspace{-8mm}
\end{figure}

\subsection{Decentralized Policy}\label{policy}
After the learning process finishes, we can evaluate how much contribution local observations $o_i$ have on $B_j$'s motors. Formally, for branches $B_i$ and $B_j (1\leq i\leq n, 1\leq j\leq n)$, \color{black} we calculate the connection strength $C_{ij}$ with L-P norm:
% \begin{equation}\label{connection}
%     C_{ij} = \frac{1}{|D|}\sum_{o_i \sim D} \frac{\sum_{m \in \mathbb{M}_j}(\pi_{\theta_i}(o_i)[m]-0)^2}{\sum_{m \in \mathbb{M}_i}(\pi_{\theta_i}(o_i)[m]-0)^2}
% \end{equation}
\begin{equation}\label{connection}
    C_{ij} = \frac{1}{|D|}\sum_{o_i \sim D}\left[\sum_{m \in \mathbb{M}_j}\left(\pi_{\theta_i}(o_i)[m]-0\right)^p\right]^{1/p}
\end{equation}\color{black}
Specifically, $C_{jj}$ represents $B_j$'s contributions to itself. If connections between $B_i$ and $B_j$ are essential to overall performance, the reinforcement learning objective will keep $C_{ij}/C_{jj}$ at high levels. If the relative connection strength $C_{ij}/C_{jj}$ is smaller than a small threshold: $C_{ij}/C_{jj} < \eta$, we can remove the connection between $B_i$ and $B_j$:
\begin{equation}\label{decouple}
    \pi_{\theta_i}(o_i)[m] =0,\quad m\in \mathbb{M}_j,\quad if \ \frac{C_{ij}}{C_{jj}} < \eta
\end{equation}

\color{black}
In addition, it is possible to implement decentralization at the motor level, particularly in cases where branches overlap. From another perspective, each motor receives influences from all branches. We can calculate the connection strength $S_{ij}$ between branch $B_i$ and an individual motor $m_j$, and then remove weak connections: 
\begin{equation}\label{connection2}
    S_{ij} = \frac{1}{|D|}\sum_{o_i \sim D}\left|\pi_{\theta_i}(o_i)[m_j]\right|
    \vspace{-2mm}
\end{equation}
\vspace{-2mm}
\begin{equation}\label{decouple}
    \pi_{\theta_i}(o_i)[m] =0,\quad m\in \mathbb{M}_j,\quad if \ \frac{S_{ij}}{S_{j}} < \eta'
\end{equation}

Here, $S_j$ represents the connection strength from the motor's own branches: $S_{j} = \sum_{m_j\in B_i}S_{ij}$.

After we perform decentralization, only essential connections remained while others are removed. 
Our approach ensures that all policies are optimized jointly and coordinated under the same central clock. As a result, each local observation has the potential to contribute to global motor actions and all policies can work together under equivalence by preserving essential connections. For instance, if a particular branch detects a collision, it can transmit signals to other branches and modify their original actions to prepare for the collision.
% In practice, we may encounter a situation where all connections can not be removed. Under this situation, we can still find our decentralized policy to be more robust (Sec. \ref{exp}). 
% After we perform decentralization, only essential connections remained while others are removed. We obtain a partially decentralized policy as shown in Figure \ref{figmethods}. The remained connections may be essential to the whole body performances so we can not remove them. In practice, we may encounter a situation where all connections can not be removed. Under this situation, we can still find our decentralized policy to be more robust (Sec. \ref{exp}). 
\color{black}
% However, our algorithm encourages branches to influence others as little as possible, we can still find our decentralized policy to be more robust to observation noises in Sec. \ref{exp}.

% \subsection{decentralized Motor Skill Learning}\label{overall_algo}

\begin{figure*}
\begin{center}
\includegraphics[width=0.5\linewidth]{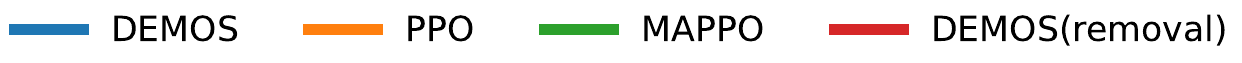}

\subfigure[Original locomotion tasks on terrains]{
\raisebox{0.4\height}{\includegraphics[width=0.28\linewidth]{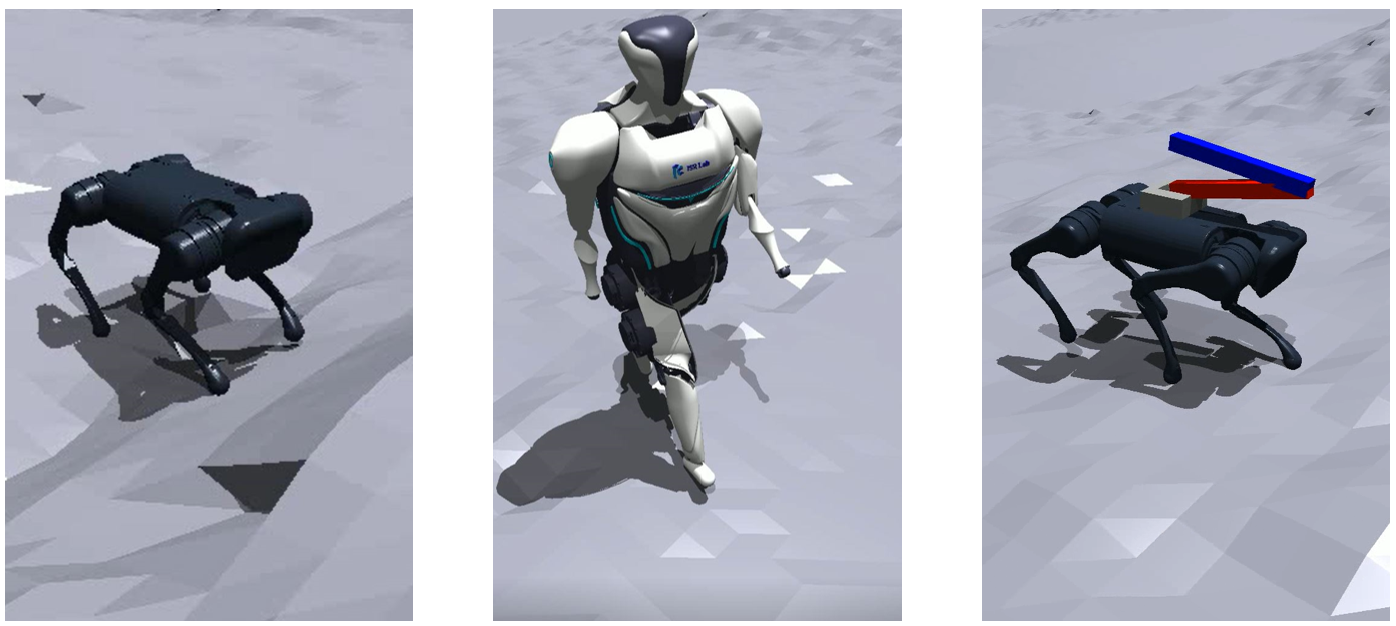}\vspace{0pt}}}
\subfigure[Quadruped locomotion]{
\includegraphics[width=0.22\linewidth]{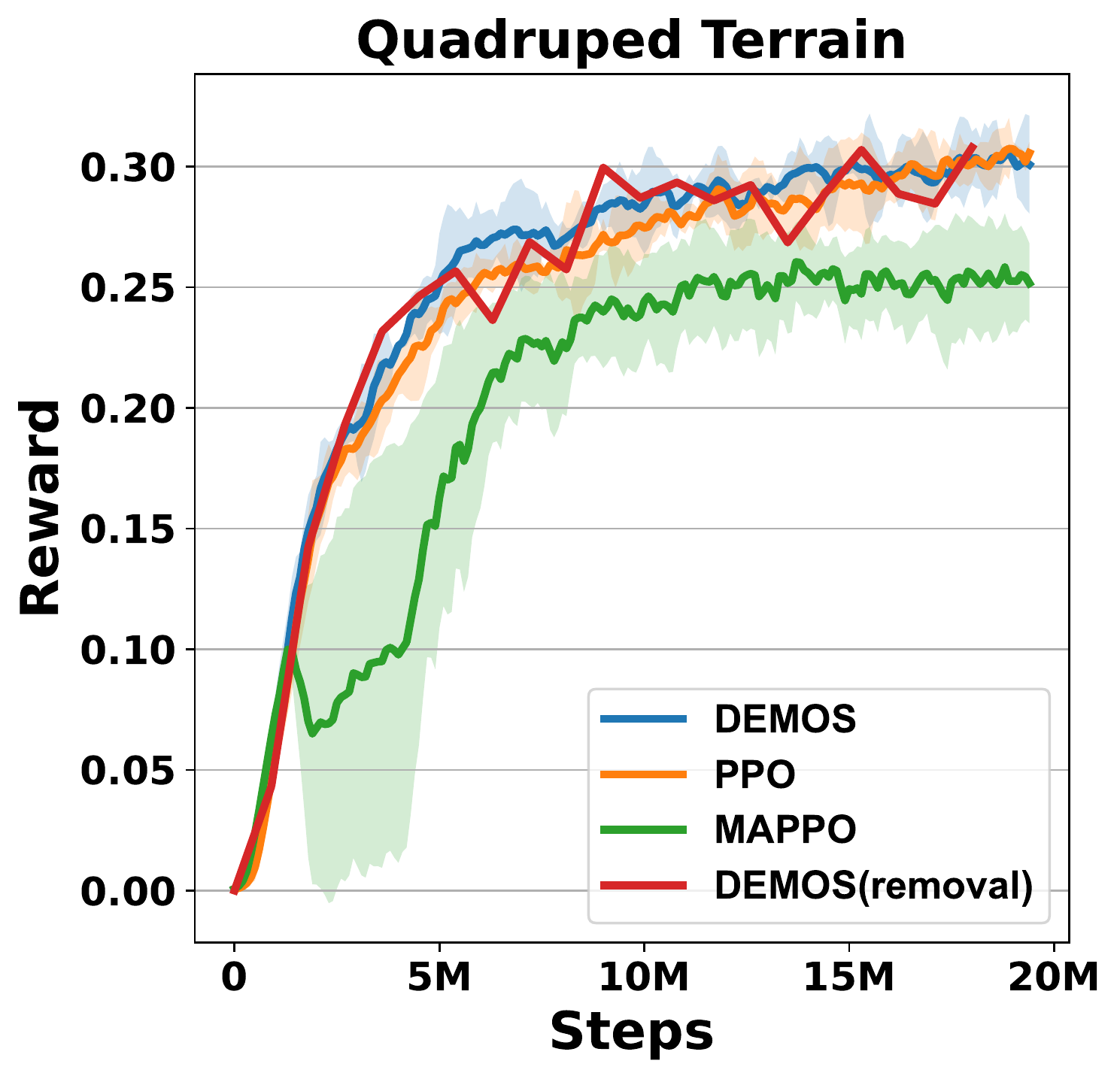}\vspace{0pt}}
\subfigure[Humanoid locomotion]{
\includegraphics[width=0.22\linewidth]{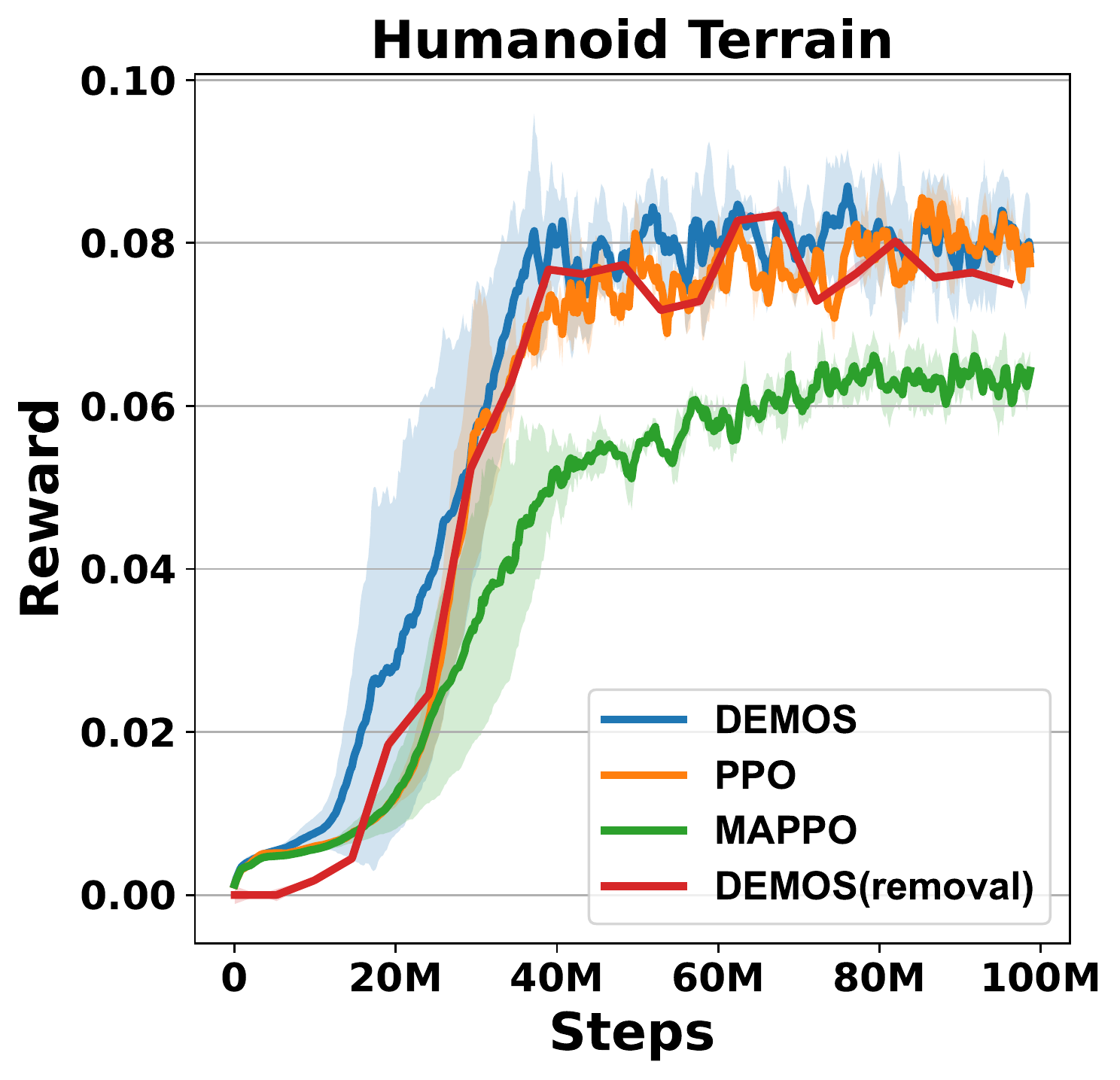}\vspace{0pt}}
\subfigure[Quadruped+Arm locomotion]{
\includegraphics[width=0.22\linewidth]{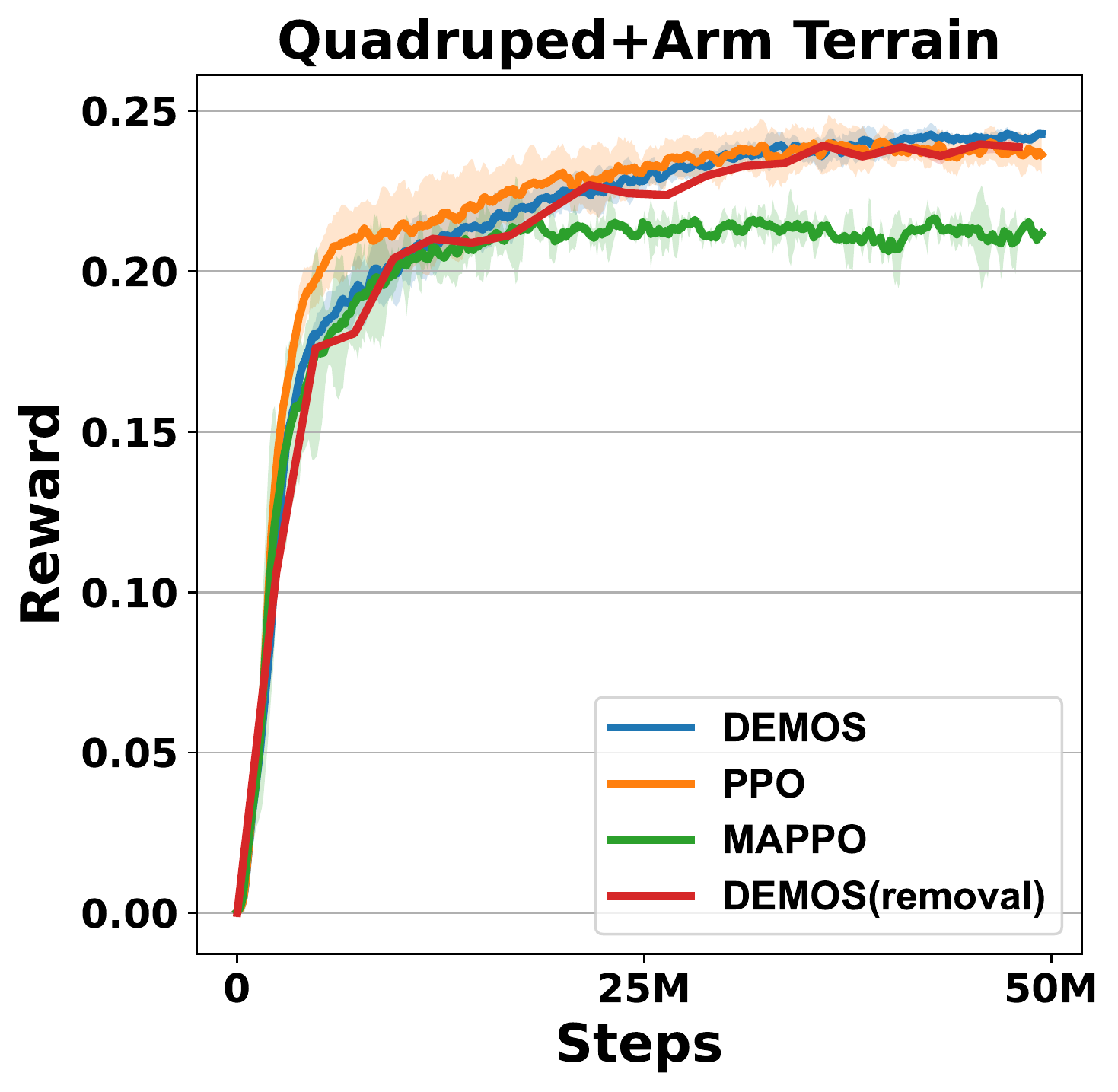}\vspace{0pt}}
\caption{\color{black}Locomotion tasks on challenging terrain for various types of robot. X-axis stands for environment interaction steps and Y-axis stands for the average reward. DEMOS(removal) denote performance after removing the weak connections.\color{black}}
\label{originaltask}
\end{center}
\vspace{-3mm}
\end{figure*}

\begin{figure*}
\begin{center}
\includegraphics[width=0.7\linewidth]{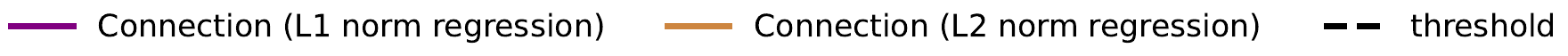}\vspace{0pt}

\subfigure[Quadruped branch connections]{
\includegraphics[width=0.32\linewidth]{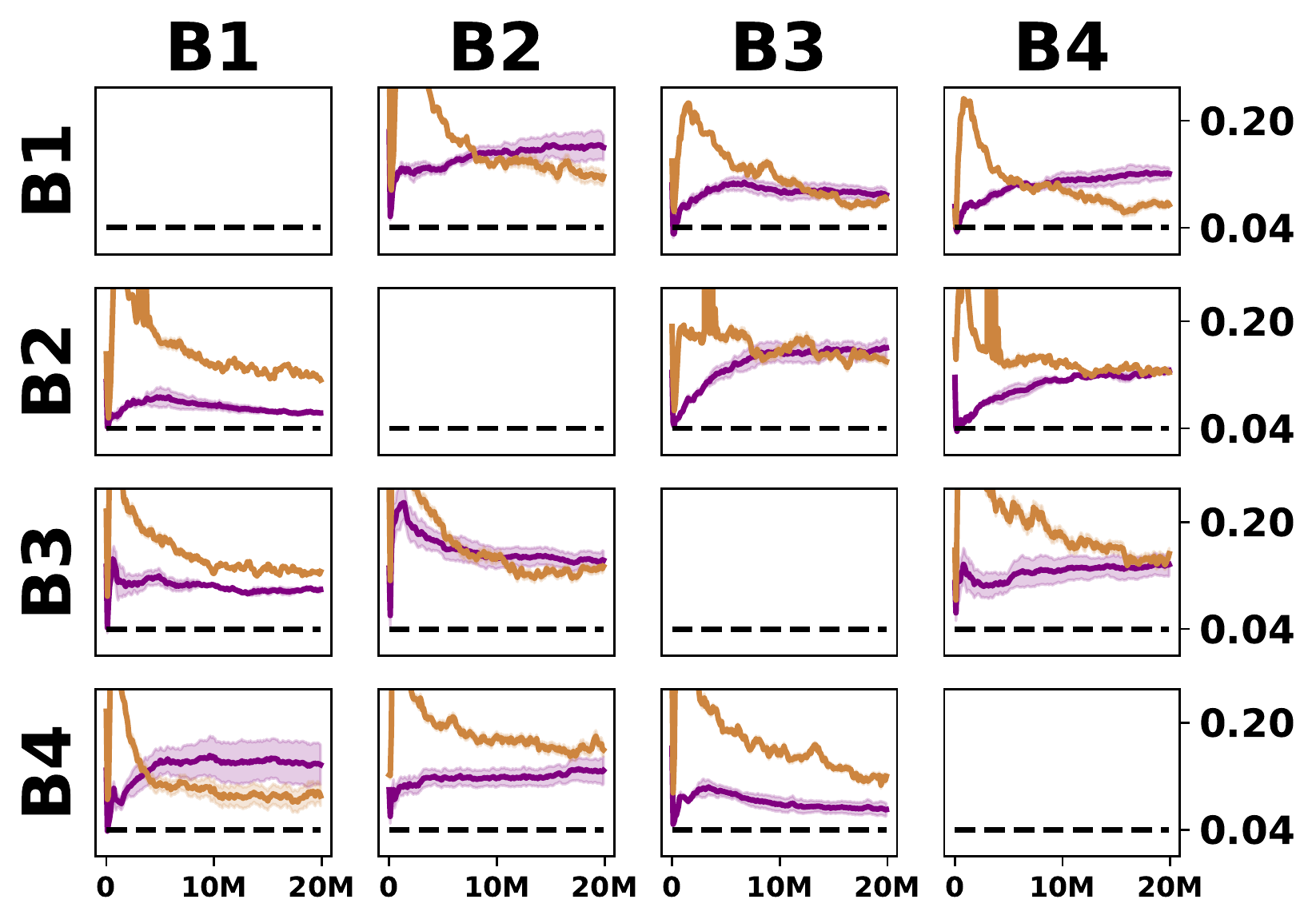}\vspace{0pt}}
\subfigure[Humanoid branch connections]{
\includegraphics[width=0.32\linewidth]{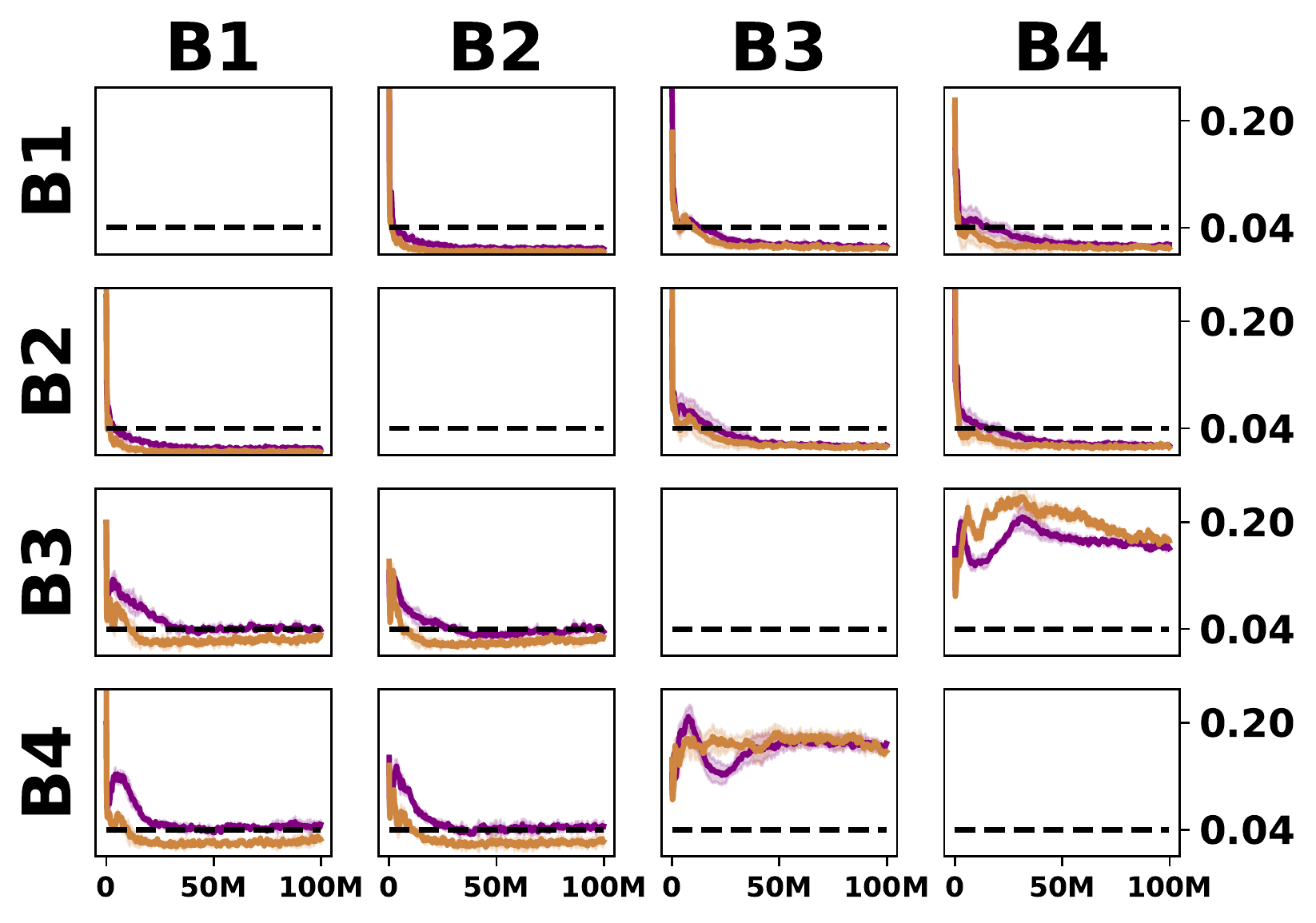}\vspace{0pt}}
\subfigure[Quadruped+Arm branch connections]{
\includegraphics[width=0.32\linewidth]{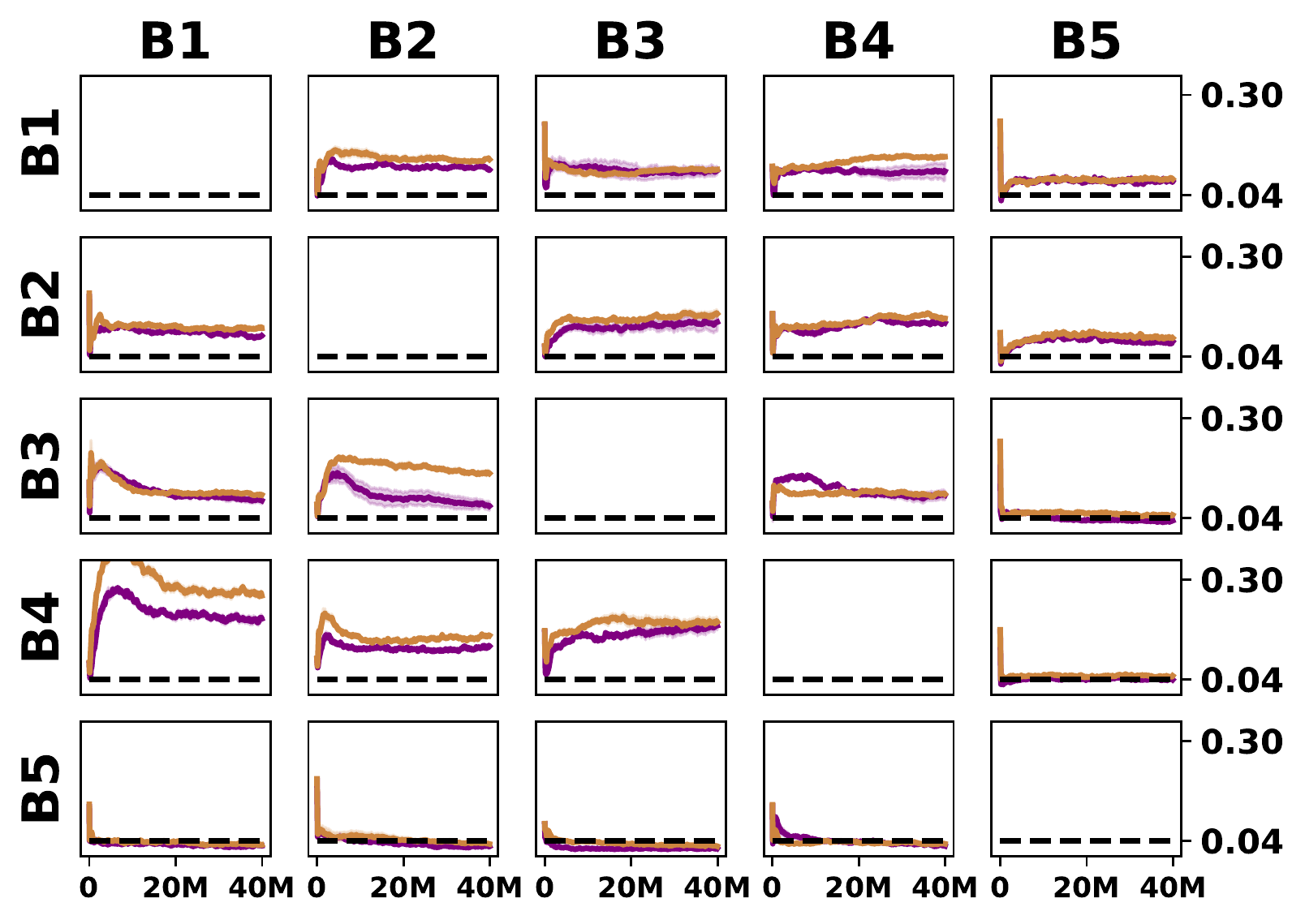}\vspace{0pt}}
% \subfigure[Performance on new tasks]{
% \includegraphics[width=0.45\linewidth]{figures/bias2_wide.pdf}\vspace{0pt}}
% \subfigure[Performance on new tasks]{
% \includegraphics[width=0.45\linewidth]{figures/bias2_wide.pdf}\vspace{0pt}}
\caption{\color{black}Visualization of normalized connection strength between branches. X-axis stands for the environment interaction step and Y-axis stands for the relative connection strength $C_{ij}/C_{jj}$. Subfigure $(i,j)$ stands for influence from $B_i$ to $B_j$ during the training, and the threshold is set to 0.04. Connections in subfigure $(i,i)$ are always equal to 1 and we do not plot it.\color{black}} 
\label{visual}
\end{center}
\vspace{-3mm}
\end{figure*}
\color{black}

\section{EXPERIMENTS}\label{exp}

In this section, we present experimental results aimed at addressing the following questions:
\begin{enumerate}
    \item Does decentralized policy sacrifice performance on the original learning task when compared to a centralized policy? (Sec. \ref{secperformance})
    \item Is decentralized policy more robust to local motor malfunctions? (Sec. \ref{secmotor})
    \item How can we transfer a decentralized policy to new tasks and rapidly acquire new skills? (Sec. \ref{secnewtask})
\end{enumerate}
\subsection{Setups and baselines}
\color{black}
We choose different types of robots in our experiments, which include a quadruped robot with 12 motors, a humanoid robot with 16 motors, and a quadruped with an attached arm, powered by 15 motors. The complete set of observation dimensions can be found in Table \ref{tabledim}.\color{black}
Our neural network policy output the PD target of all motors and runs at 50 Hz, and the low-level PD controllers run at 1000Hz with $k_p=40$ and $k_d=1$. 
% \begin{figure}[H]
%     \centering
%     \includegraphics[width=0.47\textwidth]{figures/robots.png}
%     \caption{Robots and terrains in our experiments.}
%     \label{figrobot}
% \end{figure}

% \vspace{-2mm}
\begin{table}[ht]
\centering
\small
\begin{tabular}{cccc}
\toprule
\multirow{2}{*}{} & \multirow{2}{*}{Quadruped} & \multirow{2}{*}{Humanoid} & \color{black}\multirow{2}{*}{\begin{tabular}[c]{@{}c@{}}Quadruped\\ +Arm\end{tabular}} \color{black}\\
 &  &  &  \\\midrule
Projection of gravity & 3         & 3   &3     \\
Clock inputs   & 2         & 2     &2   \\
Joint positions        & 12        & 16   &15    \\
Joint velocities    & 12        & 16     &15  \\
Last actions       & 12        & 16     &15  \\
\midrule
Overall           & 41        & 53     &50  \\
\bottomrule
\end{tabular}
\caption{Observation space for the various types of robots used in our experiments. The projection of gravity (in robot root coordinates) reflects the orientation of the root link. Clock inputs are periodic sin and cos signals used for periodic walking.}
\label{tabledim}
% \vspace{-5mm}
\end{table}

The proposed decentralized motor skill learning pipeline is compatible with various on-policy or off-policy reinforcement learning algorithms and we choose PPO \cite{schulman2017proximal} as our backbone RL algorithm. However, training robots with manually designed rewards can lead to unnatural behaviors. To mitigate this, we integrated animal and human motion capture (Mocap) data into the learning process. This Mocap data was used as a form of regularization to produce natural robot behaviors. The overall reward $r$ is the combination of task reward $r_{task}$ and style reward $r_{style}$ similar to Deepmimic \cite{peng2018deepmimic}: $r =r_{task}+r_{style}$.

\begin{figure*}[ht]
\begin{center}
\includegraphics[width=0.4\linewidth]{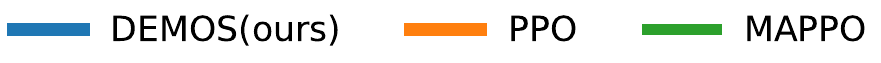}

\raisebox{0.2\height}{\includegraphics[width=0.12\linewidth]{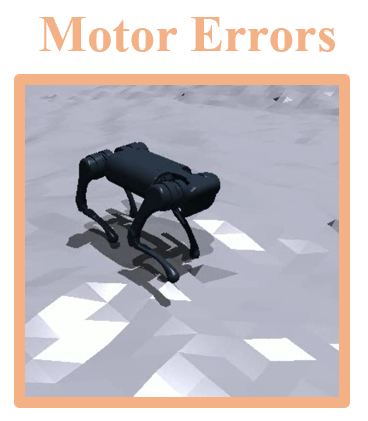}\vspace{0pt}}
\subfigure[Quadruped - Motor noise]{
\includegraphics[width=0.42\linewidth]{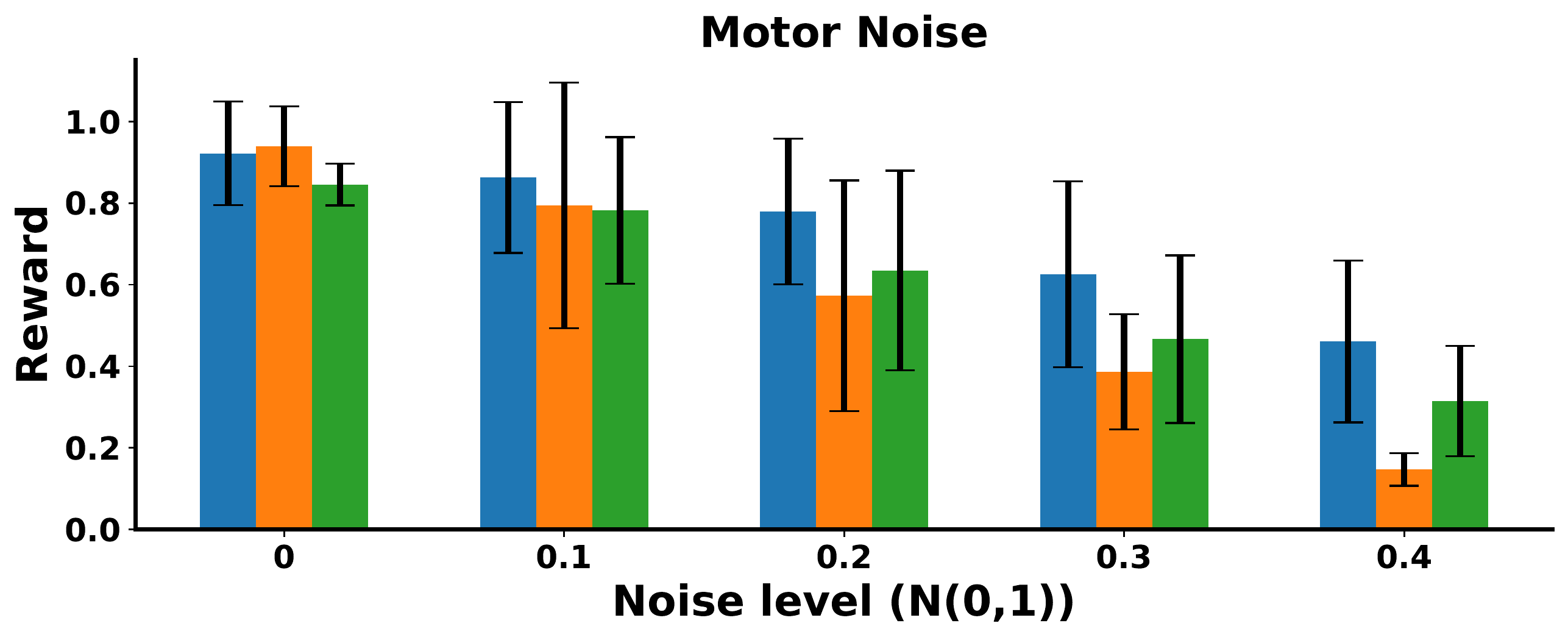}\vspace{0pt}}
\subfigure[Quadruped - Motor stuck]{
\includegraphics[width=0.42\linewidth]{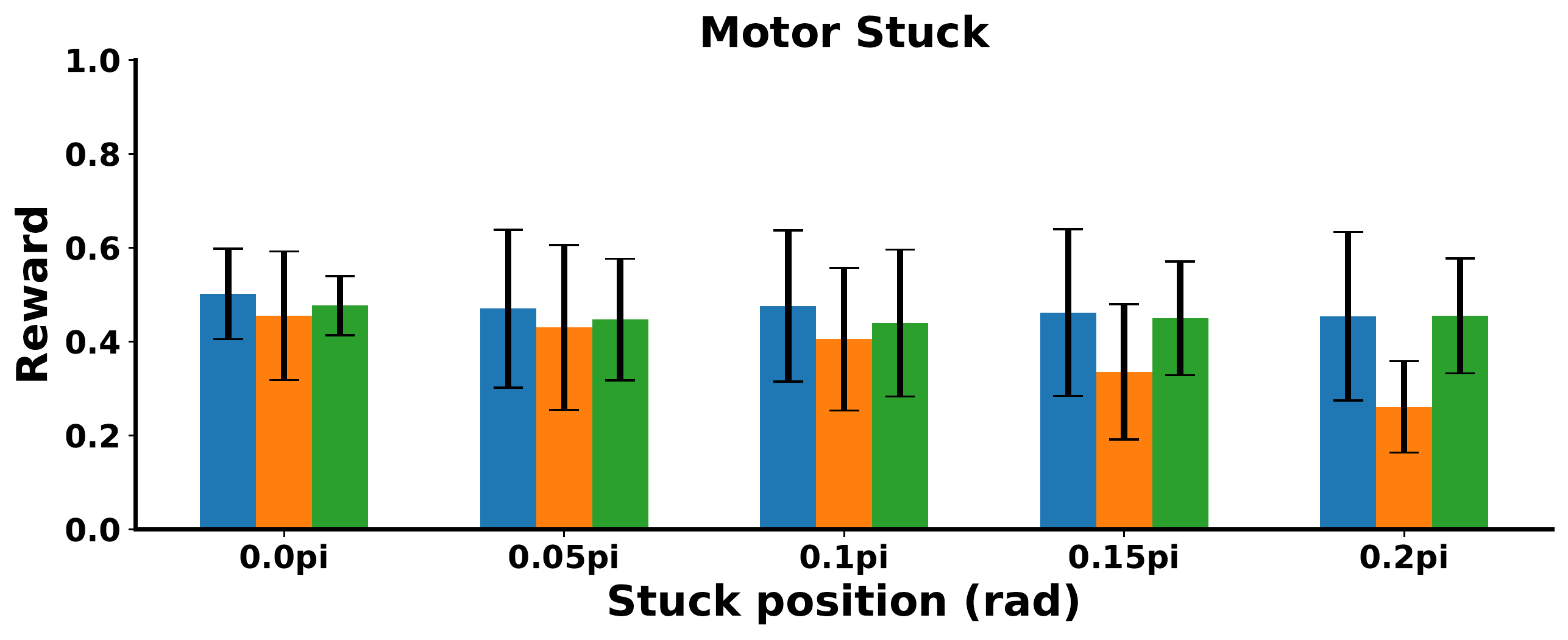}\vspace{0pt}}
\raisebox{0.2\height}{\includegraphics[width=0.12\linewidth]{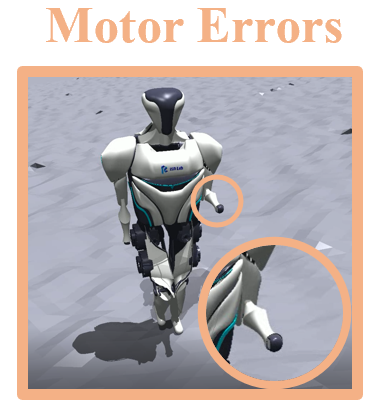}\vspace{0pt}}
\subfigure[Humanoid - Motor noise]{
\includegraphics[width=0.42\linewidth]{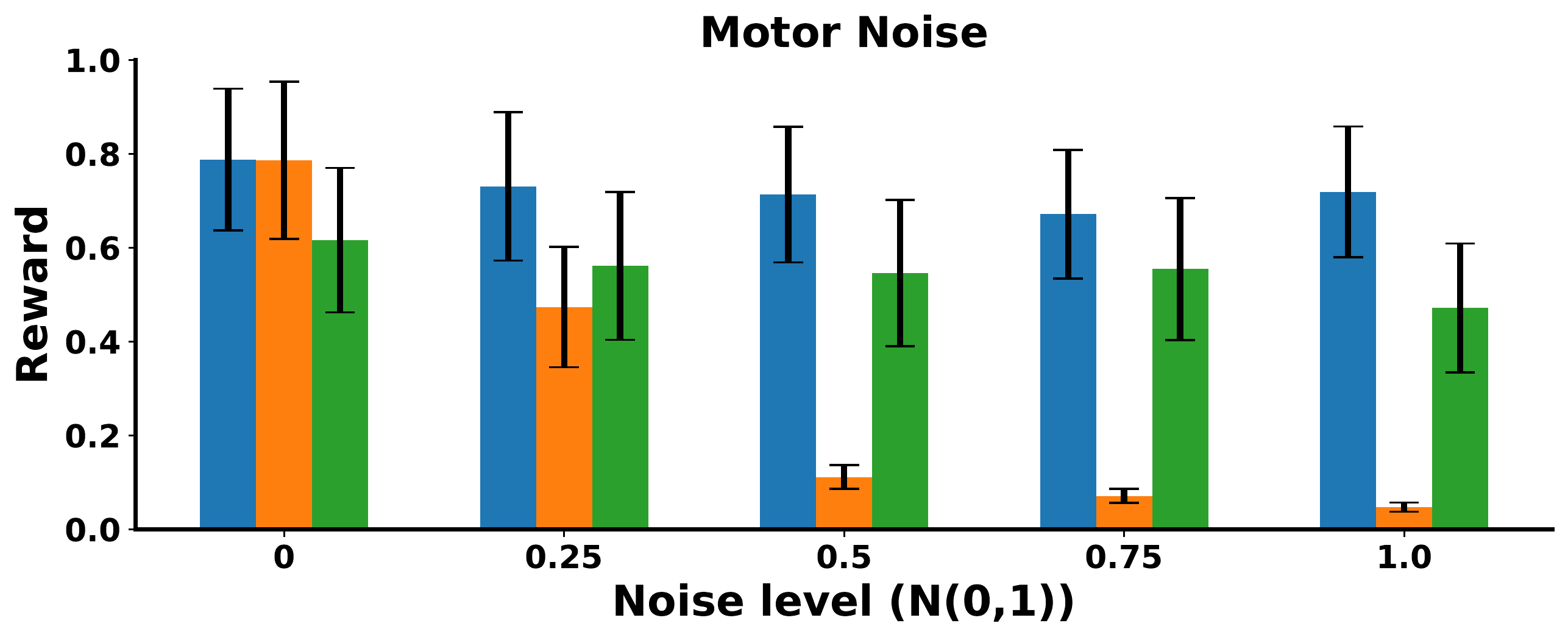}\vspace{0pt}}
\subfigure[Humanoid - Motor stuck]{
\includegraphics[width=0.42\linewidth]{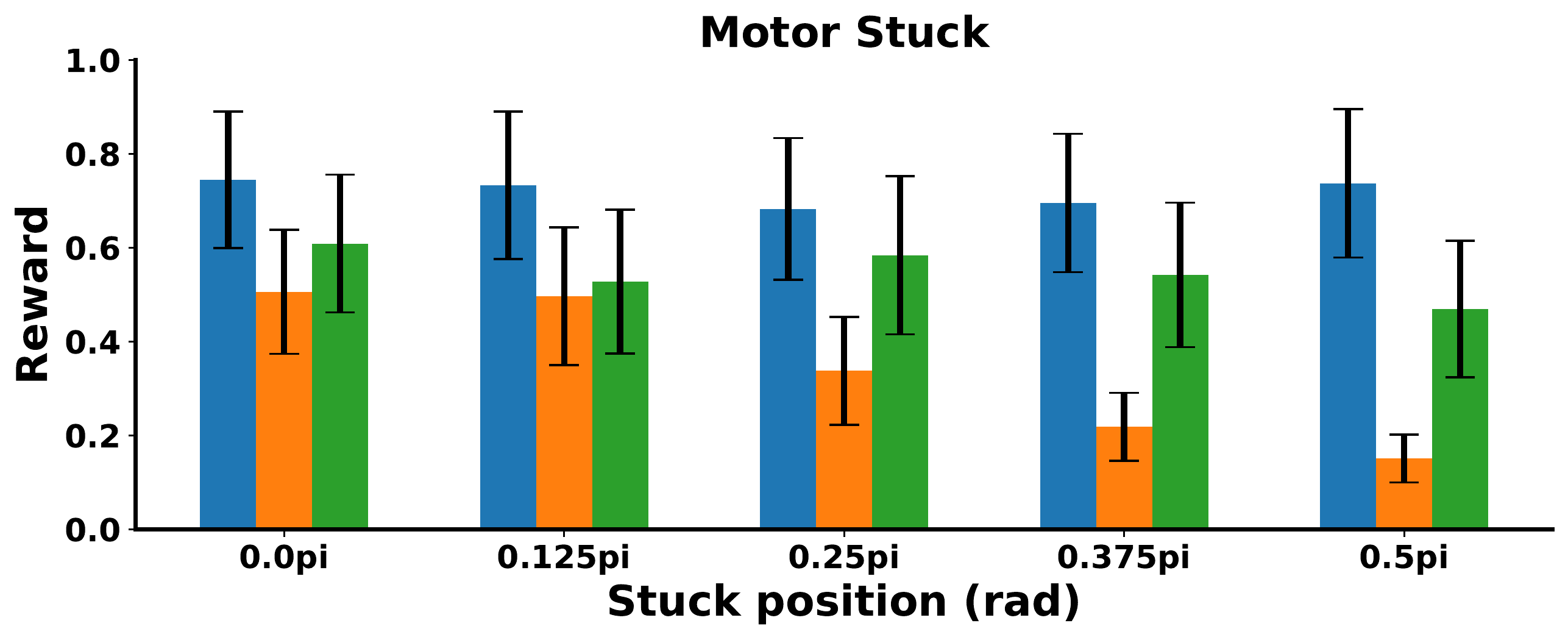}\vspace{0pt}}
\raisebox{0.2\height}{\includegraphics[width=0.12\linewidth]{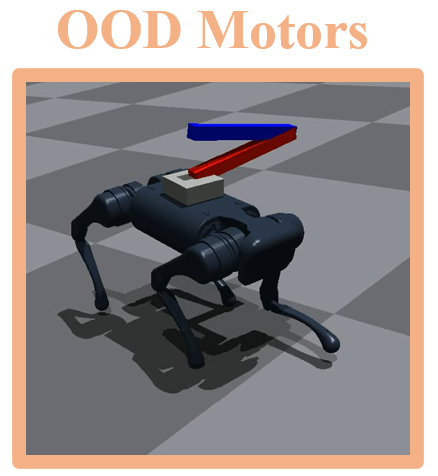}}\subfigure[Quadruped+arm - Motor noise]{
\includegraphics[width=0.42\linewidth]{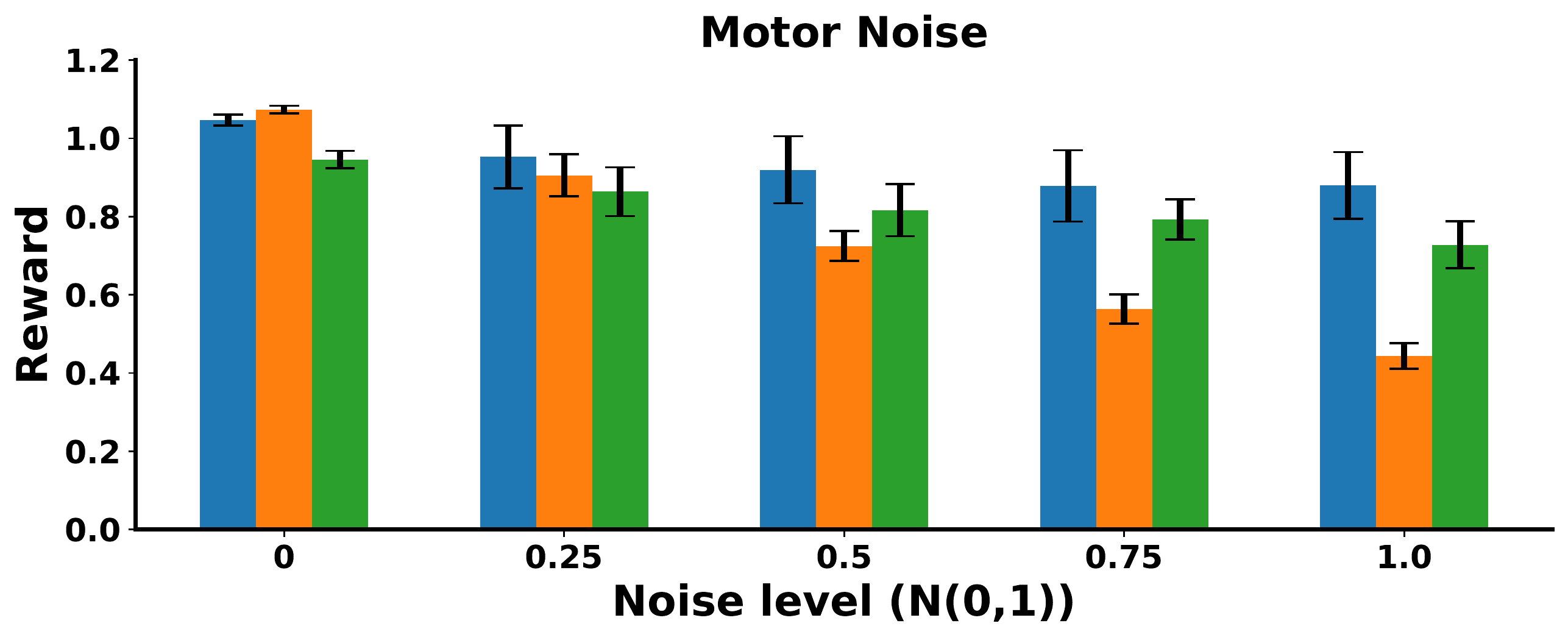}\vspace{0pt}}
\subfigure[Quadruped+arm - Motor stuck]{
\includegraphics[width=0.42\linewidth]{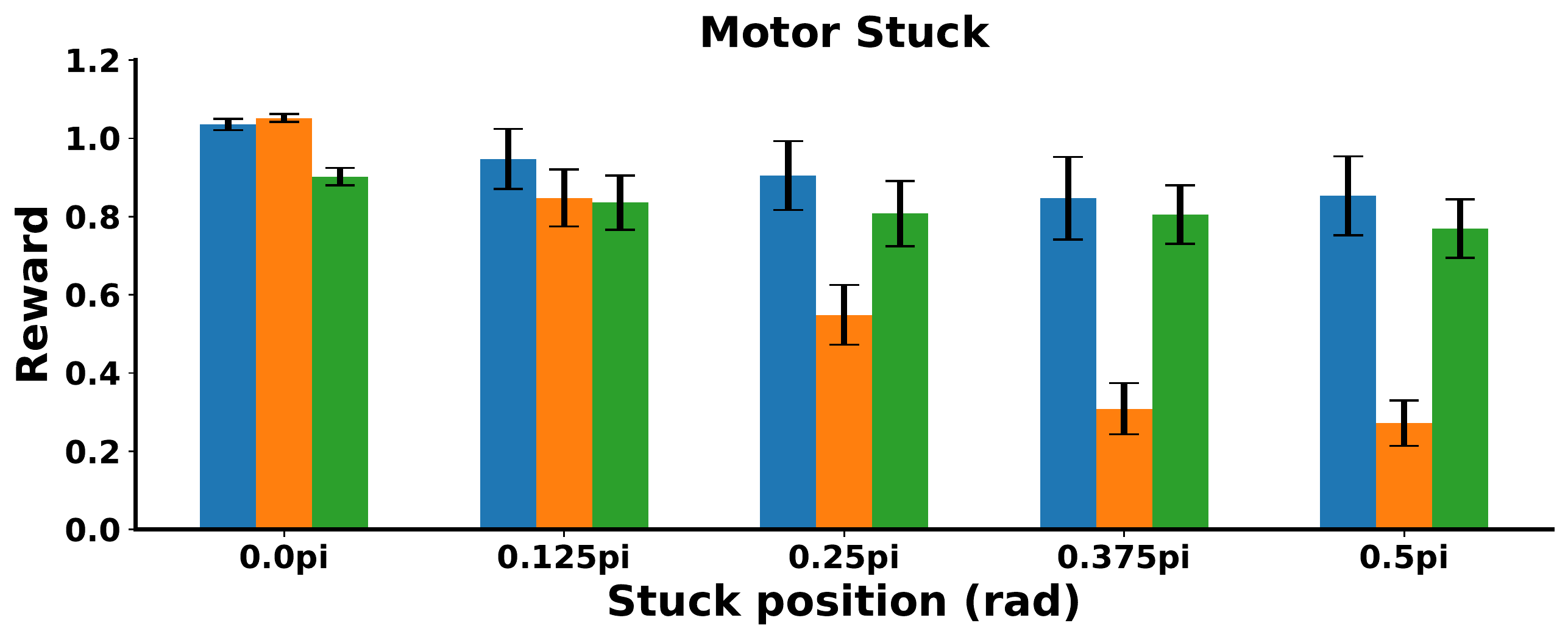}\vspace{0pt}}
\caption{\color{black}Evaluations on various types of robots with motor observation noises and motor stuck errors. We can find that local motor errors have less influence on decentralized policy since DEMOS minimizes the influences between branches.\color{black}}
\label{robust}
\end{center}
\vspace{-8mm}
\end{figure*}

Besides our proposed \textbf{De}centralized \textbf{mo}tor \textbf{s}kill learning (\textbf{DEMOS}) Method, we compared our algorithms to the centralized approach and the multi-agent approach:
\begin{itemize}
    \item \textbf{PPO \cite{schulman2017proximal}}: The most common approach in the previous works, typically using a single neural network as a centralized controller.
    \item \textbf{MAPPO \cite{yu2021surprising}}: A popular multi-agent reinforcement learning algorithm shown to have state-of-the-art performance in cooperation games. Under this setting, each module has its own actor that observes locally and acts locally.
\end{itemize}

Our simulated environment is built with Isaac Gym \cite{makoviychuk2021isaac}. All the experiments are run over 5 random seeds.

\subsection{Performance on original task}\label{secperformance}
In this part, we want to demonstrate that our proposed DEMOS algorithm will not sacrifice performance on original learning tasks compared to the centralized approach.
The quadruped robot and humanoid robot are required to walk on various types of terrains such as up-slope, down-slope, and random up-down terrains. Before the training stage,
the quadruped robot is divided into 4 branches $\{B_1,B_2,B_3,B_4\}$ which stand for \emph{\{Left-forward branch, Right-forward branch, Left-behind branch, Right-behind branch\}}. And humanoids robot is divided into 4 branches $\{B_1,B_2,B_3,B_4\}$ stand for \emph{\{Left-arm branch, Right-arm branch, Left-leg branch, Right-leg branch\}}. \color{black}Quadruped with arm is divided into 5 branches $\{B_1,B_2,B_3,B_4,B_5\}$ which stand for \emph{\{Left-forward branch, Right-forward branch, Left-behind branch, Right-behind branch, Arm branch\}}.\color{black}

% Then we initialize a local-input-local-output MLP policy for each branch. 
% We compare the learning performance of DEMOS, MLP and MAPPO.
% For the proposed DEMOS algorithm, quadruped and humanoid robots are divided into 4 branches and initialized 4 separate decentralized policies. MLP actor is initialized with a single centralized policy. MAPPO actor is initialized with 4 local-input-local-output policies.

Comparisons can be found in Figure \ref{originaltask}. We notice that our decentralized policy achieves comparable performance with a single centralized PPO policy. However, the multi-agent approach reduces the performance since each agent only contributes to its own motors, making the cooperation between modules more difficult. \color{black} Moreover, removing weak connections did not sacrifice performance, denoted as DEMOS(removal).\color{black}

% \begin{figure}[h]
% \vspace{-3mm}
% \begin{center}
% % \includegraphics[width=\linewidth]{figure/legend.pdf}
% % \includegraphics[width=0.12\linewidth]{figures/robot_error1.png}\vspace{0pt}
% \subfigure[Quadruped locomotion]{
% \includegraphics[width=0.46\linewidth]{figures/figure_a1.pdf}\vspace{0pt}}
% \subfigure[Humanoid locomotion]{
% \includegraphics[width=0.46\linewidth]{figures/figure_hum.pdf}\vspace{0pt}}
% \caption{Locomotion task on challenging terrain for the quadruped robot and the humanoid robot. X-axis stands for learning epochs and Y-axis is the average reward.}
% \label{originaltask}
% \end{center}
% \vspace{-4mm}
% \end{figure}

We also visualize the relative connection strength $C_{ij}/C_{jj}$ between modules during the DEMOS training in Figure \ref{visual}, calculated by equation \ref{connection}. 
For the quadruped locomotion task, we find that all connections between modules are relatively high, and DEMOS maintains all connections between the four branches. This means that the four branches must work in cooperation with each other to achieve optimal performance. On the other hand, in the humanoid locomotion tasks, only $B_3$ (\emph{left-foot branch}) and $B_4$ (\emph{right-foot branch}) preserve relatively high connections. DEMOS has reduced all other connections without sacrificing performance. \color{black}For the quadruped robot with an arm, we can remove the connection from the arm to the legs and preserve others.\color{black}
% \begin{figure}[h]
% \vspace{-3mm}
% \begin{center}
% % \includegraphics[width=\linewidth]{figure/legend.pdf}
% % \includegraphics[width=0.12\linewidth]{figures/robot_error1.png}\vspace{0pt}
% \subfigure{
% \includegraphics[width=0.65\linewidth]{figures/bias_legend.pdf}\vspace{0pt}}
% \vspace{-2mm}
% \subfigure[Quadruped locomotion]{
% \includegraphics[width=0.46\linewidth]{figures/figure_bias.pdf}\vspace{0pt}}
% \subfigure[Humanoid locomotion]{
% \includegraphics[width=0.46\linewidth]{figures/bias2.pdf}\vspace{0pt}}
% \caption{Visualization of connections between branches. Subfigure$(i,j)$ stands for influence from $B_i$ to $B_j$ during the training. X-axis stands for the learning step and Y-axis stands for connection strength.}
% \label{visual}
% \end{center}
% \vspace{-6mm}
% \end{figure}

\begin{figure*}
\begin{center}
\includegraphics[width=0.4\linewidth]{figures/legend.pdf}

\subfigure[New tasks]{
\raisebox{0.0\height}{\includegraphics[width=0.6\linewidth]{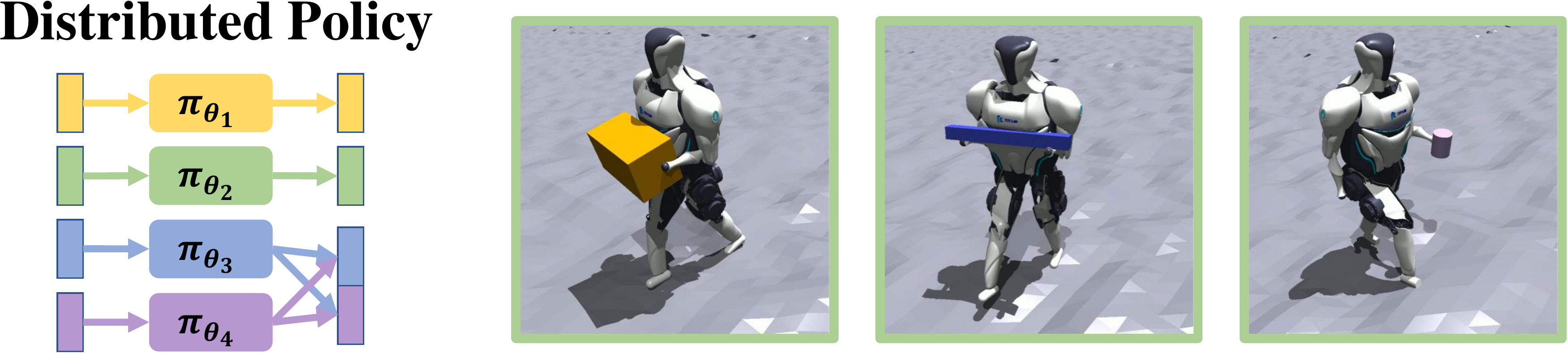}\vspace{0pt}}}
\subfigure[Performance on new tasks]{
\includegraphics[width=0.35\linewidth]{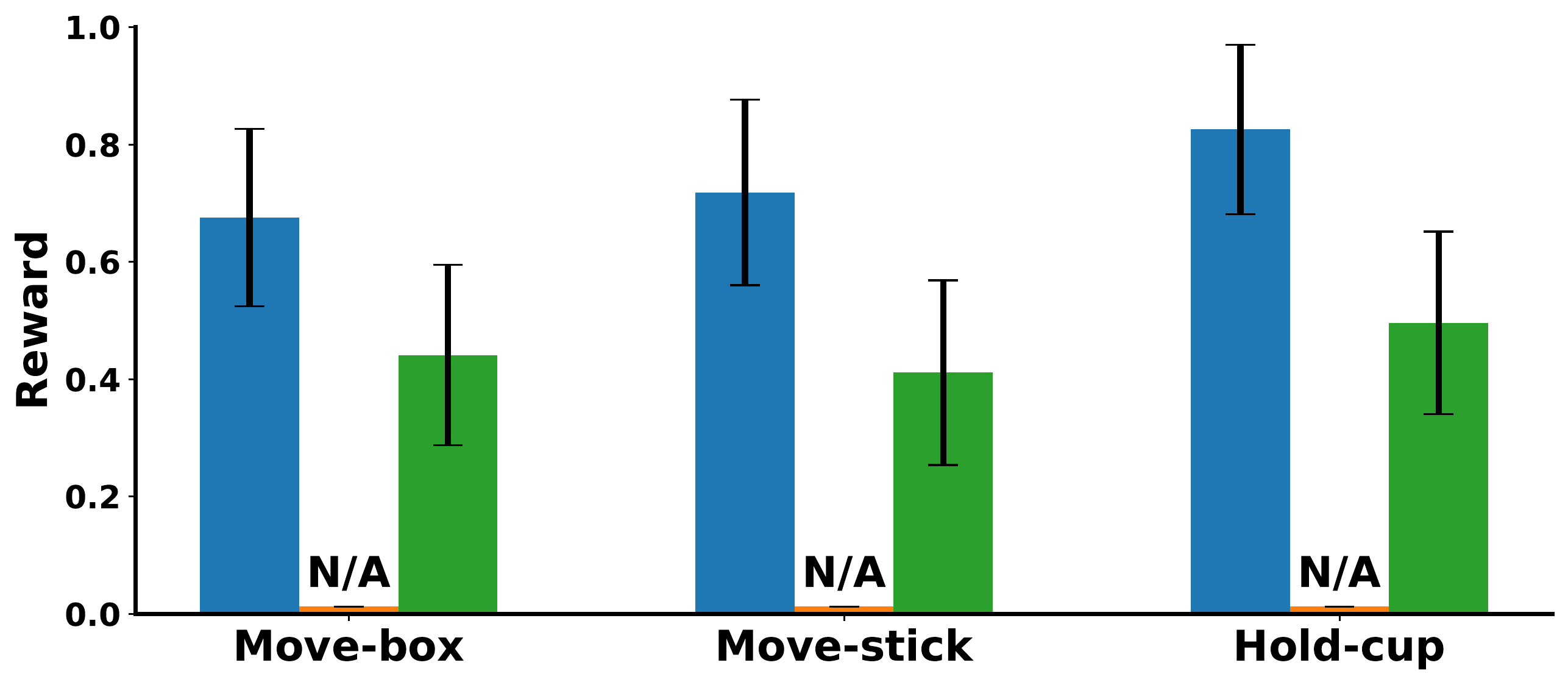}\vspace{0pt}}
\caption{Transfer part of the decentralized policy to new tasks: move-box task, move-stick task, hold-cup task.}
\label{newtask}
\end{center}
\vspace{-3mm}
\end{figure*}

\subsection{Local motor malfunctions}\label{secmotor}
In this part, we demonstrate that DEMOS learns policies that are more robust to motor malfunctions. DEMOS achieves this by encouraging each branch to have minimal influence on others. As a result, errors on a single branch may also impact other branches less. Additionally, the DEMOS pipeline may decouple the connections between two branches, which means that errors in one branch will not affect the other branch at all.

We consider 2 kinds of malfunctions that widely exist in motors:
\begin{itemize}
    \item \textbf{Motor noise}: a damaged motor may have a large motor observation noise and lead to out-of-distribution local observations.
    % We consider the situation where one robot motor is damaged and has a large observation noise. We gradually increase the noise level and check the robustness of control policies.
    \item \textbf{Motor stuck}: a motor may lose the power to move or be stuck at a certain degree by external force or obstacle. In this setting, we assume that one motor is stuck at a certain degree and cannot move.
\end{itemize}

We assume that one motor on the quadruped left-forward leg and one motor on the humanoid left arm suffer these 2 types of motor malfunctions with different levels.

Experimental results are shown in Figure \ref{robust}.  
% The decentralized policy can work against local motor malfunctions since we minimize the connections between branches, which results in less performance drop on quadruped robots compare to the centralized policy. Moreover, for the humanoid robot, motor malfunctions in the left arm almost do not harm the locomotion performance of the decentralized policy. This is reasonable since DEMOS has decoupled the arm branch and the leg branch thus errors do not propagate to the legs at all. However, the centralized PPO policy loses performance quickly since the local motor errors propagate to global motors through a centralized neural network policy.
The decentralized policy is more robust against local motor malfunctions since DEMOS minimizes the connections between branches, resulting in less performance drop for quadruped robots compared to the centralized policy. Furthermore, in the case of humanoid robots, motor malfunctions in the left arm have almost no effect on locomotion performance. This is reasonable since DEMOS has decoupled the arm branch from the leg branch, thus preventing errors from propagating to the legs. However, the centralized PPO policy experiences a rapid drop in performance as local motor error can impact global motors through a centralized neural network policy.

% It is also worth mentioning, the decentralized policy cannot resist large motor errors in the humanoid leg. This is reasonable since even humans can hardly walk with a broken leg. However, like human beings, the humanoid robot now can walk with a broken arm thanks to the decentralized policies.
It is worth mentioning that decentralized policies are incapable of resisting large motor errors in the humanoid leg. This is reasonable since even humans struggle to walk with a broken leg. However, similar to human beings, humanoid robots are now capable of walking with a broken arm, thanks to the decentralized policies.

\subsection{Transfer decentralized policy to new tasks}\label{secnewtask}
% In this part, we demonstrate that we can transfer part of the decentralized policy to new tasks and obtain new skills quickly. 
In this section, we demonstrate that it is possible to transfer subsets of the decentralized policy to new tasks and acquire new skills rapidly.

In the original task, we divide the humanoid robot into 4 branches $\{B_1,B_2,B_3,B_4\}$, and initialize policies $\pi_{\theta_1},\pi_{\theta_2},\pi_{\theta_3},\pi_{\theta_4}$ respectively. During the training stage, branches are decoupled except for the \emph{left-leg branch} and \emph{right-leg branch}. This means we can divide policies into 3 subsets: 
\begin{equation}
    \{\pi_{\theta_1}\}, \{\pi_{\theta_2}\}, \{\pi_{\theta_3},\pi_{\theta_4}\}
\end{equation}
Connections between these subsets have vanished, and each subset can function independently. Specifically, $\{\pi_{\theta_1}\}$ controls left arm to swing naturally, $\{\pi_{\theta_2}\}$ controls right arm to swing naturally and $\{\pi_{\theta_3}, \pi_{\theta_4}\}$ jointly controls 2 legs to walk on terrains. \color{black}

% When faced with new tasks, there are two approaches in which we can reuse certain sets of sub-policies.

% Firstly, we can combine them to learn new skills. For example, in the move-box task, we can utilize the leg-walking policy set ${\pi_{\theta_3}, \pi_{\theta_4}}$ that was previously used. By replacing the original arm swing policy ${\pi_{\theta_1}, \pi_{\theta_2}}$ with the grasp policy $\pi_{box}$, we can successfully accomplish the move-box task. The new sub-policy $\pi_{box}$ can be either a model-based policy or a pre-trained neural network policy.

% Secondly, these policies provide an excellent starting point for learning new tasks. In the move-box scenario, we can initiate a new policy for the arm while retaining the policy parameters in the set ${\pi_{\theta_3}, \pi_{\theta_4}}$. This way, leveraging an agent who already possesses the knowledge of walking can serve as a solid foundation for the move-box task.

When faced with new tasks, we can reuse certain sets of sub-policies in 2 approaches: (1) Combination lead to new skills. For instance, in the move-box task, we can utilize the leg-walking policy set $\{{\pi_{\theta_3}, \pi_{\theta_4}\}}$ from the previous task. To complete the move-box task, we only need to substitute the original arm swing policy ${\pi_{\theta_1}, \pi_{\theta_2}}$ with the grasp policy $\pi_{box}$. The new sub-policy $\pi_{box}$ can be either a model-based policy or a pre-trained neural network policy. (2) Provide a great starting point to learn new tasks. In the move-box case, we can initiate a new policy for the arm while retaining or freezing the policy parameters in the set ${\{\pi_{\theta_3}, \pi_{\theta_4}\}}$. Utilizing an agent who already possesses the knowledge of walking can serve as a strong foundation for the move-box task.

% In the face of new tasks, we can reuse some sub-policy sets. For instance, we can reuse leg-walking policy $\{\pi_{\theta_3}, \pi_{\theta_4}\}$ in the move-box task. We can just replace the original arm swing policy $\pi_{\theta_1},\pi_{\theta_2}$ with grasp policy $\pi_{box}$ to finish move-box task. The new sub-policy $\pi_{box}$ can either be a model-based policy or a pre-trained neural network policy. We can also initiate a new policy for the arm while keeping the policy parameters in the set $\{\pi_{\theta_3}, \pi_{\theta_4}\}$ to learn new skills. An agent who already knows how to walk will be a good starting point for move-box task. 
In our experiments, we combine pre-designed model-based arm policies and certain previously learned policies to handle new tasks:\color{black}
\begin{itemize}
    \item Move-box skill: $\{\pi_{box}\} + \{\pi_{\theta_3},\pi_{\theta_4}\}$
    \item Move-stick skill: $\{\pi_{stick}\} + \{\pi_{\theta_3},\pi_{\theta_4}\}$
    \item Hold-cup skill: $\{\pi_{cup}\}+ \{\pi_{\theta_2}\} + \{\pi_{\theta_3},\pi_{\theta_4}\}$
\end{itemize}

Here $\pi_{box}$ is a model-based controller for 2 arms to hold a box. $\pi_{stick}$ is a controller for 2 arms to hold a long stick. $\pi_{cup}$ is a controller for a single arm to hold a cup. 

Experimental results are shown in Figure \ref{newtask}. The decentralized policy can be successfully transferred to new tasks and combined into new skills. However, a typical centralized PPO policy is highly task-specific with all modules coupled together, thus can not transfer to new tasks. MAPPO policies lose performance when transferred to new tasks since some important cooperation between modules has been destroyed.

\section{CONCLUSIONS}
In this work, we propose decentralized Motor skill learning (DEMOS), a decentralized approach for complex robotic systems control. Instead of one typical neural network policy that is highly centralized, we leverage multiple local-input-global-output policies to control the robot and decouple modules with weak connections automatically during training. 
Experiment results demonstrated that DEMOS are effective for a wide range of robot types, including quadrupeds, humanoids, and quadrupeds with arms, etc. \color{black}However, In the face of robots with largely overlapped branch structure, we may need other division rules in the pre-training stage to better decentralize the control policy.\color{black}
% Experiments on quadruped and humanoid robots demonstrate that our method improves the robustness and generalization of the policy without sacrificing performance.
% will not sacrifice performance but can be more robust to local motor errors. Moreover, We can even transfer subsets of the decentralized policies to new tasks and obtain new skills quickly.

% \addtolength{\textheight}{-8cm}   
% This command serves to balance the column lengths
                                  % on the last page of the document manually. It shortens
                                  % the textheight of the last page by a suitable amount.
                                  % This command does not take effect until the next page
                                  % so it should come on the page before the last. Make
                                  % sure that you do not shorten the textheight too much.

%%%%%%%%%%%%%%%%%%%%%%%%%%%%%%%%%%%%%%%%%%%%%%%%%%%%%%%%%%%%%%%%%%%%%%%%%%%%%%%%

%%%%%%%%%%%%%%%%%%%%%%%%%%%%%%%%%%%%%%%%%%%%%%%%%%%%%%%%%%%%%%%%%%%%%%%%%%%%%%%%

%%%%%%%%%%%%%%%%%%%%%%%%%%%%%%%%%%%%%%%%%%%%%%%%%%%%%%%%%%%%%%%%%%%%%%%%%%%%%%%%
\section*{APPENDIX}

\subsection{Implementation details}
Our simulation environments are built in IsaacGym \cite{makoviychuk2021isaac} and we implement our DEMOS algorithm and baselines in Pytorch based on opensource codebase https://github.com/leggedrobotics/legged\_gym \cite{rudin2022learning}.

Hyperparameters of the backbone PPO algorithm can be found in table \ref{parameters}.

% A comparison of neural network actors can be found in table \ref{actor}, DEMOS and MAPPO have multiple actors and hold small networks for each actor.  
\begin{table}[H]
\centering
\begin{tabular}{cc}
\toprule
                  Parameters & Value \\ 
\midrule
Number of Environments         & 4096        \\
Learning epochs           & 5        \\
Steps per Environment           & 24       \\
Minibatch Size &24576\\
\color{black}Episode length\color{black}          & 20 seconds       \\
Discount Factor             & 0.99       \\
Generalised Advantage Estimation(GAE)             & 0.95       \\
PPO clip         & 0.2       \\
Entropy coefficient           & 0.005        \\
Desired KL           & 0.01        \\
Learning Rate          & 5e-4       \\
Weight decay            & 0.01       \\
\bottomrule
\end{tabular}
\caption{Hyperparameters of backbone PPO algorithm.}
\label{parameters}
\vspace{-3mm}
\end{table}

% \begin{table}[H]
% \centering
% \begin{tabular}{ccc}
% \toprule
%                   Algorithm & Hidden layers & Activation \\ 
% \midrule
% DEMOS(ours)         & [256,128,64]  &Elu      \\
% MLP(PPO)           & [512,256,128]    &Elu    \\
% MAPPO            & [256,128,64]    &Elu   \\
% \bottomrule
% \end{tabular}
% \caption{Neural network size for each policy.}
% \label{actor}
% % \vspace{-5mm}
% \end{table}
% \section*{ACKNOWLEDGMENT}
% This work is supported by the Ministry of Science and Technology of the People’s Republic of China, the 2030 Innovation Megaprojects “Program on New Generation Artificial Intelligence” (Grant No. 2021AAA0150000).

%%%%%%%%%%%%%%%%%%%%%%%%%%%%%%%%%%%%%%%%%%%%%%%%%%%%%%%%%%%%%%%%%%%%%%%%%%%%%%%%
\bibliographystyle{IEEEtran}
\bibliography{IEEEexample}

\end{document}